%% file: camera-ready.tex
\pdfoutput=1

\documentclass[11pt]{article}

\usepackage[]{acl}

\usepackage{times}
\usepackage{latexsym}

\usepackage[T1]{fontenc}

\usepackage[utf8]{inputenc}

\usepackage{microtype}
\usepackage{bm}
\usepackage{amsmath}
\usepackage{graphics}
\usepackage{graphicx}
\usepackage{caption}
\usepackage{subcaption}
\usepackage{multirow}
\usepackage{adjustbox}
\usepackage{pgfplots}
\usepackage{tikz}
\usepackage{pgf-pie}
\usepackage{diagbox}
\usepackage[multiple]{footmisc}
\usepackage{bbm}
\usepackage{booktabs}

\title{Bidirectional Transformer Reranker for Grammatical Error Correction}

\author{Ying Zhang$^{1}$, Hidetaka Kamigaito$^{2}$, and Manabu Okumura$^{1, 3}$ \\
  $^{1}$Tokyo Institute of Technology \\
  $^{2}$NARA Institute of Science and Technology  \\
  $^{3}$RIKEN Center for Advanced Intelligence Project \\
  \texttt{\{zhang,oku\}@lr.pi.titech.ac.jp \quad kamigaito.h@is.naist.jp} \\
  }

\begin{document}
\maketitle
\begin{abstract}
\textcolor{black}{Pre-trained seq2seq models have achieved state-of-the-art results in the grammatical error correction task.}
\textcolor{black}{However, these models still suffer from a prediction bias due to their unidirectional decoding.}
Thus, we propose a bidirectional Transformer reranker (BTR), that re-estimates the probability of each candidate sentence generated by the pre-trained seq2seq model.
The BTR preserves the seq2seq-style Transformer architecture but utilizes a BERT-style self-attention mechanism in the decoder to compute the probability of each target token by using masked language modeling to capture bidirectional representations from the target context. For guiding the reranking, the BTR adopts negative sampling in the objective function to minimize the unlikelihood. During inference, the BTR gives final results after comparing the reranked top-1 results with the original ones by an acceptance threshold $\lambda$.
Experimental results show that, in reranking candidates from a pre-trained seq2seq model, T5-base, the BTR on top of T5-base could yield 65.47 and 71.27 $F_{0.5}$ scores on the CoNLL-14 and BEA test sets, respectively, and yield 59.52 GLEU score on the JFLEG corpus, with improvements of 0.36, 0.76 and 0.48 points compared with the original T5-base.
Furthermore, when reranking candidates from T5-large, the BTR on top of T5-base improved the original T5-large by 0.26 points on the BEA test set.\footnote{\textcolor{black}{Our code is available at \url{https://github.com/zhangying9128/BTR}.}}
\end{abstract}

\section{Introduction}
\begin{table*}[ht]
\centering
\resizebox{0.8\linewidth}{!}{
\begin{tabular}{ll}
\hline
Source & \textcolor{blue}{Speed camera} can be placed in many locations along a highway . \\
Gold 1 & \textcolor{blue}{Speed cameras} can be placed in many locations along a highway . \\
Candidate 1 (RoBERTa, T5GEC) & \textcolor{blue}{A speed camera} can be placed in many locations along a highway . \\
Candidate 2 (BTR, R2L) & \textcolor{blue}{Speed cameras} can be placed in many locations along a highway .\\
Candidate 3 & \textcolor{blue}{A Speed camera} can be placed in many locations along a highway . \\ \hline
\end{tabular}
}
\caption{Examples of reranked outputs from the JFLEG \citep{napoles-etal-2017-jfleg} test set. The 3 candidate sentences were generated by T5GEC (\S \ref{sec:compared_methods}). \textcolor{blue}{Blue} indicates the range of corrections. ``Candidate 1 (T5GEC)'' denotes that T5GEC regards ``Candidate 1'' as the most grammatical correction.}
\label{table:example_of_reranked_outputs_main}
\end{table*}

Grammatical error correction (GEC) is a sequence-to-sequence task which requires a model to aim to correct an ungrammatical sentence. \textcolor{black}{An example is presented in Table \ref{table:example_of_reranked_outputs_main}}.\footnote{\textcolor{black}{Appendix \ref{sec:appendix_example_of_reranked_outputs} presents more examples from the CoNLL-14 \citep{ng2014conll} and JFLEG test sets.}}
Various neural models for GEC have emerged \citep{Junczys-Dowmunt18,kiyono-etal-2019-empirical,kaneko-etal-2020-encoder,rothe-etal-2021-simple} due to the importance of this task for language-learners who tend to produce ungrammatical sentences.

Previous studies have shown that GEC can be approached as machine translation by using a seq2seq model \citep{luong-etal-2015-effective} with a Transformer \citep{NIPS2017_3f5ee243} architecture \citep{Junczys-Dowmunt18,zhao-etal-2019-improving, kiyono-etal-2019-empirical, kaneko-etal-2020-encoder,rothe-etal-2021-simple}.
As a neural model consists of an encoder and a decoder, \textcolor{black}{the seq2seq} architecture typically requires a large amount of training data.
Because GEC suffers from limited training data, applying a seq2seq model for GEC results in a low-resource \textcolor{black}{setting}, that can be handled by introducing synthetic data for training \citep{kiyono-etal-2019-empirical,omelianchuk-etal-2020-gector, stahlberg-kumar-2021-synthetic}. 
\textcolor{black}{However, as pointed out by \citet{rothe-etal-2021-simple}, the use of synthetic data in GEC may result in a distributional shift and require language-specific tuning, which can be time-consuming and resource-intensive.}

Considering the limitations of the synthetic data, the current trend is to utilize the learned and general representations from a pre-trained model, such as BERT \citep{devlin-etal-2019-bert}, XLNet \citep{yang2019xlnet}, BART \citep{lewis2020bart}, and T5 \citep{2020t5}, 
\textcolor{black}{which have been trained on large corpora and shown to be effective for various downstream tasks.}
\textcolor{black}{According to \citet{kaneko-etal-2020-encoder}, incorporating a pre-trained masked language model (MLM) into a seq2seq model could facilitate correction.}
In addition, as reported by \citet{rothe-etal-2021-simple}, 
\textcolor{black}{the pre-trained T5 model achieved state-of-the-art results on GEC benchmarks for four languages after successive fine-tuning with the cleaned LANG-8 corpus (cLang-8) \citep{rothe-etal-2021-simple}.}

Although the seq2seq model with pre-trained representations has shown to be effective for GEC, its performance was still constrained by its unidirectional decoding. As suggested by \citet{liu-etal-2021-neural}, for an ungrammatical sentence, a fully pre-trained seq2seq GEC model \citep{kiyono-etal-2019-empirical} could generate several high-quality grammatical sentences by beam search. 
\textcolor{black}{However, even among these candidates, there may be still a gap between the selected hypothesis and the most grammatical one.}
Our experimental results\textcolor{black}{,} listed in Table \ref{table:main_result}\textcolor{black}{,} also demonstrate their investigation.
To solve this decoding problem, given the hypotheses of a seq2seq GEC model, \citet{kaneko-etal-2019-tmu} used BERT to classify between ungrammatical and grammatical hypotheses, and reranked them on the basis of the classification results.
The previous studies \citep{kiyono-etal-2019-empirical,kaneko-etal-2020-encoder} also showed that the seq2seq GEC model decoding in an opposite direction, i.e., right-to-left, is effective as a reranker for a left-to-right GEC model. 

\textcolor{black}{Therefore, to further improve the performance of the pre-trained seq2seq model for GEC, it is essential to find ways to leverage the bidirectional representations of the target context.} 
In this study, on the basis of the seq2seq-style Transformer model, we propose a bidirectional Transformer reranker (BTR) to handle the interaction between the source sentence and the bidirectional target context.
The BTR utilizes a BERT-style self-attention mechanism in the decoder to predict each target token by masked language modeling \citep{devlin-etal-2019-bert}.
\textcolor{black}{Given} several candidate target sentences from a base model, the BTR can re-estimate the sentence probability for each candidate from the bidirectional representation of the candidate, which is different from the conventional seq2seq model.
During training, for guiding the reranking, we adopt negative sampling \citep{NIPS2013_9aa42b31} for the objective function to minimize the unlikelihood while maximizing the likelihood. In inference, considering the robustness of pre-trained models, we compare the reranked top-1 results with the original ones by an acceptance threshold $\lambda$ to decide whether to accept the suggestion from the BTR.

We regard the state-of-the-art model for GEC \citep{rothe-etal-2021-simple}, 
\textcolor{black}{a pre-trained Transformer model, T5 (either T5-base or T5-large),} as our base model and utilize its generated candidates for reranking. 
Because the BTR can inherit learned representations from a pre-trained Transformer model, we construct the BTR on top of T5-base. 
Our experimental results showed that, by reranking candidates from a fully pre-trained and fine-tuned T5-base model, the BTR on top of T5-base can achieve an $F_{0.5}$ score of 65.47 on the CoNLL-14 benchmark.
\textcolor{black}{The BTR on top of T5-base also outperformed T5-base on the BEA test set by 0.76 points, achieving an $F_{0.5}$ score of 71.27.} 
Adopting negative sampling for the BTR also generated a peaked probability distribution for ranking, and so grammatical suggestions could be selected by using $\lambda$.
Furthermore, the BTR on top of T5-base was robust even when reranking candidates from T5-large and improved the performance by 0.26 points on the BEA test set.

\section{Related Work}
For directly predicting the target corrections from corresponding input tokens, \citet{omelianchuk-etal-2020-gector} and \citet{malmi-etal-2022-text} regarded the encoder of the Transformer model as a non-autoregressive GEC sequence tagger.
The experimental results of \citet{omelianchuk-etal-2020-gector} showed that, compared with the randomly initialized LSTM \citep{6795963}, the pre-trained models, such as RoBERTa \citep{liu2019roberta}, GPT-2 \citep{radford2019language}, and ALBERT \citep{lan2019albert}, can achieve higher $F_{0.5}$ scores as a tagger.
\citet{sun2021instantaneous} considered GEC as a seq2seq task and introduced the Shallow Aggressive Decoding (SAD) for the decoder of the Transformer. With the SAD, the performance of a pre-trained seq2seq model, BART, surpassed the sequence taggers of \citet{omelianchuk-etal-2020-gector}.
The T5 xxl model is a pre-trained seq2seq model with 11B parameters \citep{2020t5}. After fine-tuning with the cLang-8 corpus, T5 xxl and mT5 xxl \citep{xue-etal-2021-mt5}, a multilingual version of T5, achieved state-of-the-art results on GEC benchmarks in four languages: English, Czech, German, and Russian \citep{rothe-etal-2021-simple}.
This demonstrated that performing a single fine-tuning step for a fully pre-trained seq2seq model is a simple and effective method for GEC without incorporating a copy mechanism \citep{zhao-etal-2019-improving}, the SAD or the output from a pre-trained MLM \citep{kaneko-etal-2020-encoder}.
Despite the improvements brought about by the pre-trained representations, the conventional seq2seq structure suffers from a prediction bias due to its unidirectional decoding. According to \citet{liu-etal-2021-neural}, by using beam search, a fully pre-trained seq2seq GEC model \citep{kiyono-etal-2019-empirical} can generate several high-quality grammatical hypotheses, which include one that is more grammatical than the selected one.

To address the shortcoming of the unidirectional decoding, previous studies \citep{kiyono-etal-2019-empirical, kaneko-etal-2019-tmu, kaneko-etal-2020-encoder} introduced reversed representations to rerank the hypotheses. \citet{kiyono-etal-2019-empirical} and \citet{kaneko-etal-2020-encoder} utilized a seq2seq GEC model that decodes in the opposite direction (right-to-left) to rerank candidates, which was effective to select a more grammatical sentence than the original one. This finding motivated us to use a bidirectional decoding method for our model. 
Instead of using a seq2seq model, \citet{kaneko-etal-2019-tmu} fine-tuned BERT as a reranker to evaluate the grammatical quality of a sentence.
By using masked language modeling, BERT learned deep bidirectional representations to distinguish \textcolor{black}{between} grammatical and ungrammatical sentences. However, BERT did not account for the positions of corrections
\textcolor{black}{, as it discarded the source sentence and considered only the target sentence.}
This made it difficult for BERT, as a reranker, to recognize the most suitable corrected sentence for an ungrammatical sentence. \citet{salazar-etal-2020-masked} proposed the \textcolor{black}{use of} pseudo-log-likelihood scores ($\mathrm{PLL}$) for reranking. 
\textcolor{black}{They demonstrated that RoBERTa, with the $\mathrm{PLL}$ for reranking, outperformed the conventional language model GPT-2 when reranking candidates in speech recognition and machine translation tasks.}
\citet{zhang-etal-2021-language} also claimed that the pre-trained model, MPNet \citep{song2020mpnet}, was more effective than GPT-2 when using $\mathrm{PLL}$ for reranking in discourse segmentation and parsing.

\citet{zhang-van-genabith-2021-bidirectional} proposed a bidirectional Transformer-based alignment (BTBA) model\textcolor{black}{, which aims} to assess the alignment between the source and target tokens in machine translation.
\textcolor{black}{To achieve this,} BTBA masked and predicted the current token with attention to both left and right sides of the target context to produce alignments for the current token. 
Specifically, to assess alignments from the attention scores in all cross-attention layers, the decoder in BTBA discarded the last feed-forward layer of the Transformer model and directly predicted masked tokens from the output of the last cross-attention layer. Even though the target context on both sides was taken into consideration, one limitation of BTBA was that the computed alignments ignored the representation of the current token. 
To produce more accurate alignments, \citet{zhang-van-genabith-2021-bidirectional} introduced full context based optimization (FCBO) for fine-tuning, in which BTBA no longer masks the target sentence to use the full target context.

In our research, 
\textcolor{black}{to determine the most appropriate correction for a given erroneous sentence, we model the BTR as a seq2seq reranker, which encodes the erroneous sentence using an encoder and decodes a corrected sentence using a decoder.}
\textcolor{black}{In contrast to the conventional seq2seq model, we use masked language modeling to mask and predict each target token in the decoder and estimate the sentence probability for each candidate using $\mathrm{PLL}$.}
Unlike BTBA, the BTR preserves the last feed-forward layer in the decoder to predict masked tokens more accurately. Because the original data of the masked tokens should be invisible in the prediction, the FCBO fine-tuning step is not used in the BTR.
Compared with BTBA, the BTR keeps the structure of the Transformer model and can easily inherit parameters from pre-trained models.

\section{\textcolor{black}{Preliminary}}
\textcolor{black}{Because} the decoder of the BTR uses masked language modeling to rerank candidates based on the $\mathrm{PLL}$, 
\textcolor{black}{in this section, we explain how a Transformer-based GEC model generates the candidates}, the masked language modeling used in BERT, and how to compute the $\mathrm{PLL}$.

\subsection{Transformer-based GEC Model}
\label{seq2seq_gec}
\begin{figure}[t]
    \begin{subfigure}[t]{0.47\columnwidth}
        \includegraphics[width=0.8\columnwidth]{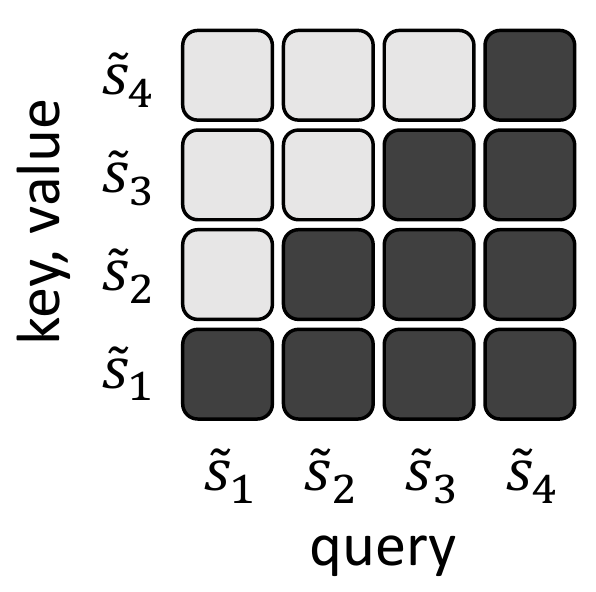}
        \caption{\textcolor{black}{Causal}}
        \label{fig:causal_attention}
    \end{subfigure}
    \hfill
    \centering
    \begin{subfigure}[t]{0.47\columnwidth}
        \includegraphics[width=0.8\columnwidth]{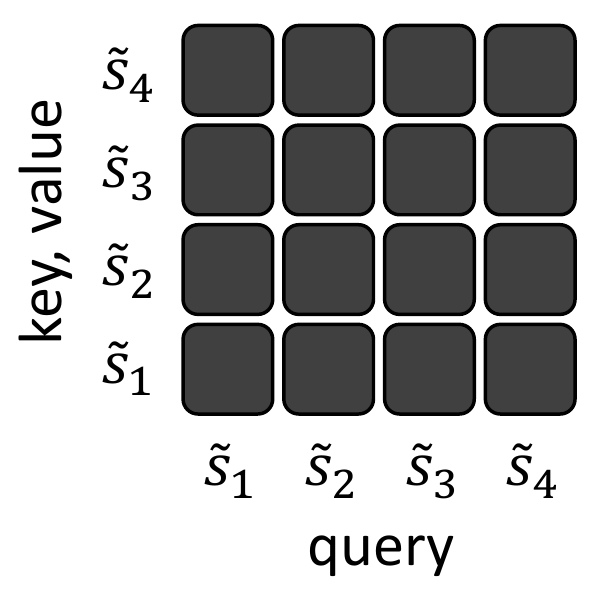}
        \caption{Fully-visible}
        \label{fig:fully_attention}
    \end{subfigure}
\caption{Mask patterns in the Transformer model \citep{NIPS2017_3f5ee243} (a) and in the BTR (b) for the self-attention mechanism in the decoder. 
Light cells indicate no attention.}
\end{figure}

Given an ungrammatical sentence $\bm{x}=(x_1, \ldots, x_n)$, a GEC model corrects $\bm{x}$ into its grammatical sentence $\bm{y} = (y_1, \ldots, y_m)$, where $x_i$ is the $i$-th token in $\bm{x}$ and $y_j$ is the $j$-th token in $\bm{y}$. 
As an auto-regressive model, a Transformer-based GEC model with parameter $\theta$ decomposes 
\textcolor{black}{$p(\bm{y}|\bm{x};\theta)$}
as follows: 
\begin{align}
&\textcolor{black}{\log p(\bm{y}|\bm{x};\theta) \!=\!\! \sum_{j=1}^{m} \log p(y_j|\bm{x}, \bm{y}_{<j};\theta), }\label{eq:auto_regressive}\\
& \textcolor{black}{p(y_j|\bm{x},\bm{y}_{<j};\theta) \!=\! \mathrm{softmax}(W\tilde{s}_{j}+b),}
\label{eq:linear_softmax}
\end{align}
where $\tilde{s}_{j}$ is the final hidden state from the decoder at the $j$-th decoding step. $W$ is a weight matrix, $b$ is a bias term, and \textcolor{black}{$\bm{y}_{<j}$ denotes $(y_1, \ldots, y_{j-1})$.}
\textcolor{black}{$\tilde{s}_{j}$ is computed as described in Appendix \ref{sec:appendix_computation_for_sj}.}

\subsection{\textcolor{black}{Decoding Method}}
\label{sec:appendix_comparison_of_diverse_decoding_methods}

\begin{table}[t]
\resizebox{\columnwidth}{!}{
\begin{tabular}{lrrr}
\toprule
\textbf{Method} & \textbf{Gold (\%)} & \textbf{Unique (\%)} & \textbf{Oracle ($F_{0.5}$)} \\ \midrule
Nucleus sampling & 28.70 & 43.61 & 49.27 \\
Top-k sampling & 29.62 & 48.57 & 48.40 \\
Beam search & \textbf{37.97} & \textbf{98.93} & \textbf{55.11} \\
Diverse beam search & 28.46 & 38.78 & 50.39\\ \bottomrule
\end{tabular}
}
\caption{\textcolor{black}{Results for the T5GEC on the CoNLL-13 corpus with various decoding methods.}}
\label{table:comparison_of_diverse_decoding_methods}
\end{table}

\textcolor{black}{The pre-trained T5 model with Transformer architecture achieved state-of-the-art results in GEC by using beam search for decoding \citep{rothe-etal-2021-simple}.}
\textcolor{black}{However, previous studies \citep{li2016mutual,vijayakumar2017diverse} have suggested that beam search tends to generate sequences with slight differences.}
\textcolor{black}{This can constrain the upper bound score when reranking candidates \citep{ippolito-etal-2019-comparison}.}
\textcolor{black}{To select the optimal decoding method for a Transformer-based GEC Model, T5GEC,} we compared beam search with diverse beam search \citep{vijayakumar2017diverse}, top-k sampling \citep{fan-etal-2018-hierarchical}, and nucleus sampling \citep{Holtzman2020The}.
For each pair of data in CoNLL-13 corpus \citep{ng-etal-2013-conll}, we required all decoding methods to generate 5 candidate sequences with a maximum sequence length of 200. When using diverse beam search, we fixed the beam group and diverse penalty to 5 and 0.4, respectively. Meanwhile, we set the top-k as 50 and the top-p as 0.95 for top-k sampling and nucleus sampling, respectively.

\textcolor{black}{Table \ref{table:comparison_of_diverse_decoding_methods} presents the compared results among different decoding methods.}
Oracle indicates the upper bound score that can be achieved with the generated candidates. If the candidates include the correct answer, we assume the prediction is correct. Unique (\%) indicates the rate of unique sequences among all candidates. Gold (\%) indicates the rate of pairs of data whose candidates include the correct answer. The results show that beam search generates more diverse sentences with the highest Oracle score compared to nucleus sampling, top-k sampling, and diverse beam search. This may be because, in the GEC task, most of the tokens in the target are the same as the source, which causes a peaked probabilities distribution to focus on one or a small number of tokens. And thus, a top-k filtering method like beam search generates more diverse sentences than sampling or using probability as a diverse penalty.
\textcolor{black}{Based on these results, we have chosen beam search as the decoding method for T5GEC during inference.} 
\textcolor{black}{For evaluating T5GEC, it generates the top-ranked hypothesis with a beam size of 5.}
\textcolor{black}{To generate the top-$a$ candidates $\mathcal{Y}_{a} = \{\bm{y}_{1}, \ldots,\bm{y}_{a} \}$ for reranking, it generates hypotheses with a beam size of $a$ and a maximum sequence length of 128 and 200 for the datasets in training and prediction, respectively.}

\subsection{Masked Language Modeling}
Masked language modeling, used in BERT, was introduced to learn bidirectional representations for a given sentence $\bm{x}$ through self-supervised learning \citep{devlin-etal-2019-bert}. Before pre-training, several tokens in $\bm{x}$ are randomly replaced with the mask token <M>.
Let $\kappa$ denote the set of masked positions, $\bm{x}_{\kappa}$ the set of masked tokens, and $\bm{x}_{\backslash \kappa }$ the sentence after masking. The model parameter $\theta$ is optimized by maximizing the following objective:
\begin{eqnarray}
\textcolor{black}{\log p(\bm{x}_{\kappa}|\bm{x}_{\backslash \kappa };\theta)\approx \sum_{k \in \kappa} \log p(x_{k}|\bm{x}_{\backslash \kappa};\theta).}
\end{eqnarray}
Similar to Eq.~(\ref{eq:linear_softmax}), a linear transformation with a softmax function is utilized for the final hidden state $\tilde{h}_{k}$ to predict 
\textcolor{black}{$p(x_{k}|\bm{x}_{\backslash \kappa };\theta)$.}
$\tilde{h}_{k}$ is computed as described in Appendix \ref{sec:appendix_computation_for_hkl}.
The corresponding $\mathrm{PLL}$ for $\bm{x}$ is computed by
\begin{equation}
\mathrm{PLL}(\bm{x};\theta) :=\sum_{i=1,\kappa=\{i\}}^{|\bm{x}|}\log p(x_i|\bm{x}_{\backslash \kappa};\theta ),
\end{equation}
\textcolor{black}{where $|\bm{x}|$ is the length of $\bm{x}$.}
As suggested by \citet{salazar-etal-2020-masked}, when using $\mathrm{PLL}$ to estimate the cross-entropy loss, the loss of $x_i|\bm{x}_{\backslash \kappa}$ versus $i$ from BERT is flatter than GPT-2, that uses the chain rule. Considering the candidate sentences might have different lengths, $\mathrm{PLL}$ is ideal for reranking.

\section{Bidirectional Transformer Reranker (BTR)}
\label{section:bidirectional_reranker}

\begin{figure}[t]
    \centering
    \includegraphics[width=1\columnwidth]{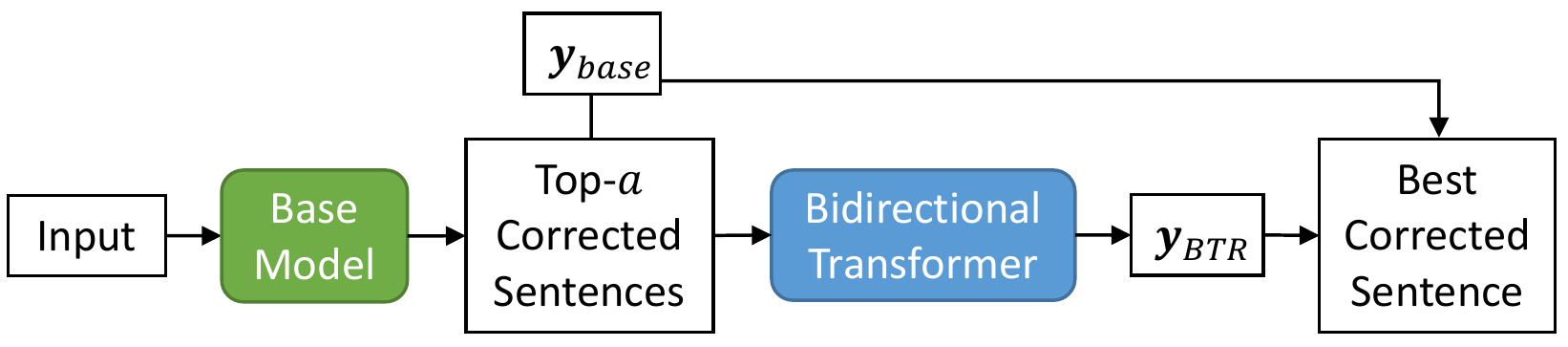}
    \caption{\textcolor{black}{Overview of the reranking procedure by using the bidirectional Transformer reranker (BTR).}}
    \label{fig:procedure}
\end{figure}

\textcolor{black}{The BTR uses masked language modeling in the decoder to estimate the probability of a corrected sentence.}
Given an ungrammatical sentence $\bm{x}$, a base GEC model first generates the top-$a$ corrected sentences \textcolor{black}{$\mathcal{Y}_{a}$}, as described in Section \ref{seq2seq_gec}. Assume $\bm{y}_{base} \textcolor{black}{\in \mathcal{Y}_{a}}$ is the top-ranked hypothesis from the base GEC model. The BTR selects and accepts the most optimal corrected sentence $\bm{y}_{BTR}$ from $\mathcal{Y}_{a}$ on the basis of the estimated sentence probability, as described in the following. Figure \ref{fig:procedure} shows the overview of the BTR for the whole procedure.

\subsection{Target Sentence Probability}
\label{section:sentence_prob}

\textcolor{black}{As $\mathrm{PLL}$ has been effective in estimating the sequence probability for reranking, we decompose the conditional sentence probability of $\bm{y}$ as:}
\begin{align}
\!\!\!&\log p(\bm{y}|\bm{x}; \theta) \nonumber\\ 
\!\!\!\approx & \mathrm{PLL}(\bm{y}|\bm{x}; \theta) = \!\!\!\!\!\!\! \sum_{j=1,\kappa=\{j\}}^{|\bm{y}|} \!\!\!\!\!\! \log p(y_j|\bm{x}, \bm{y}_{\backslash \kappa};\theta).
\end{align}
As in Eq.~(\ref{eq:linear_softmax}), a linear transformation with the softmax function is utilized for the final hidden state $\tilde{s}_{j}$ to predict $p(y_j|\bm{x}, \bm{y}_{\backslash \kappa};\theta)$. 

\textcolor{black}{Same as the Transformer architecture, $\tilde{s}_{j}$ is the result of $s_{j}$ after the cross-attention and feed-forward layers. We assume the decoder consists of $L$ layers. To capture the bidirectional representation, for $\ell \in L$, we compute $s_{j}^{\ell}$ as:}
\begin{eqnarray}
s_{j}^{\ell} &= \mathrm{Attn}_{s}(\tilde{s}_{j}^{\ell-1}, \widetilde{S}_{\backslash \kappa }^{\ell-1}, \widetilde{S}_{\backslash \kappa }^{\ell-1}),
\end{eqnarray}
where $\tilde{s}_{j}^{0}$ is the embedding of the $j-1$-th word in $\bm{y}_{\backslash \kappa}$ and $\tilde{s}_{1}$ is the state of the start token <s>. $\widetilde{S}_{\backslash \kappa }^{\ell-1}=(\tilde{s}_{1}^{\ell-1}, \ldots,\tilde{s}_{m+1}^{\ell-1})$ denotes a set of hidden states for the joint sequence of <s> and $\bm{y}_{\backslash \kappa}$. \textcolor{black}{$\mathrm{Attn}_{s}$ indicates the self-attention layer.}
Figure \ref{fig:fully_attention} shows our fully-visible attention mask for computing $S^{\ell}$ in parallel.
\textcolor{black}{The procedure of using the BTR to predict $p(y_j|\bm{x}, \bm{y}_{\backslash \kappa};\theta)$ is shown in Appendix \ref{sec:appendix_procedure}.}

\subsection{Objective Function}
\label{objective_function}
As a reranker, for a given ungrammatical sentence $\bm{x}$, the BTR should compare all corresponding corrected sentences \textcolor{black}{$\mathcal{Y}$} and select the most grammatical one. However, considering all possible corrected sentences for $\boldsymbol x$ is intractable, as suggested by \citet{stahlberg-byrne-2019-nmt}, so we consider a subset of sequences $\mathcal{Y}_{a}$ based on the top-$a$ results from the base GEC model instead.

Let $\boldsymbol y_{gold} \in \mathcal{Y}$ denote the gold correction for $\boldsymbol x$.
For $\boldsymbol y \in \mathcal{Y}_{a} \cup \{\boldsymbol y_{gold}\} $, we follow the setting of BERT to randomly mask 15\% of $\boldsymbol y$ and denote $\kappa$ as the set of masked positions. As a result, the distribution of the masked tokens satisfies the 8:1:1 masking strategy. 
Following previous research \citep{welleck-etal-2019-dialogue,zhang-etal-2021-language,song-etal-2021-bob}, given the masked sentence $\bm{y}_{\backslash \kappa }$, the model parameter $\theta$ of the BTR is optimized by maximizing the likelihood and minimizing the unlikelihood as:
\begin{align} \label{eq:maxmize_likelyhood}
&\log p(\bm{y}_{\kappa}|\bm{x},\bm{y}_{\backslash \kappa };\theta) \\ \nonumber
\approx & \sum_{k \in \kappa}[\mathbbm{1}_{\boldsymbol y}  \log p(y_k|\bm{x}, \bm{y}_{\backslash \kappa };\theta) \\ \nonumber
&+(1-\mathbbm{1}_{\boldsymbol y})\log (1 - p(y_k|\bm{x}, \bm{y}_{\backslash \kappa };\theta))],
\end{align}
where $ p(y_k|\bm{x}, \bm{y}_{\backslash \kappa };\theta)$ is computed as in Section \ref{section:sentence_prob}. $\mathbbm{1}_{\boldsymbol y}$ is an indicator function defined as follows:
\begin{eqnarray}
\mathbbm{1}_{\boldsymbol y} := \begin{cases}
1 & \text{if } \boldsymbol y=\boldsymbol y_{gold}    \\
0 & \text{if } \boldsymbol y \neq \boldsymbol y_{gold}
\end{cases}.
\end{eqnarray}

\subsection{Inference}
\label{sec:inference}
In inference, for $\boldsymbol y \in \mathcal{Y}_{a}$, the BTR scores $\bm{y}$ by
\begin{equation}
f(\bm{y}|\bm{x}) =\frac{\exp (\mathrm{PLL}(\bm{y}|\bm{x};\theta)/|\bm{y}|)}{\sum_{\bm{y}' \in \mathcal{Y}_{a}} \exp(\mathrm{PLL}(\bm{y}'|\bm{x};\theta)/|\bm{y}'|)}.
\end{equation}
Hereafter, we denote $\bm{y}_{BTR} \textcolor{black}{\in \mathcal{Y}_{a}}$ as the candidate with the highest score $f(\bm{y}_{BTR}|\bm{x})$ for given $\bm{x}$ in the BTR. Here, $f(\bm{y}|\bm{x})$ is also considered to indicate the confidence of the BTR. Because the BTR is optimized with Eq.~(\ref{eq:maxmize_likelyhood}), a high score for $\bm{y}_{BTR}$ indicates a confident prediction while a low score indicates an unconfident prediction.

Considering that we build the base GEC model from a fully pre-trained seq2seq model and the BTR from an insufficiently pre-trained model, we introduce an acceptance threshold \textcolor{black}{$\lambda$} to decide whether to accept the suggestion from the BTR.
We accept $\bm{y}_{BTR}$ only when it satisfies the following equation; otherwise, $\bm{y}_{base}$ is still the final result: 
\begin{eqnarray}
f(\bm{y}_{BTR}|\bm{x}) - f(\bm{y}_{base}|\bm{x}) > \lambda,
\label{eq:accpet_threshold}
\end{eqnarray}
where $\lambda$ is a hyperparameter tuned on the validation data. 

\section{Experiments}
\subsection{Compared Methods}
\label{sec:compared_methods}
We evaluated the BTR as a reranker for two versions of candidates, normal and high-quality ones, generated by two seq2seq GEC models, T5GEC and T5GEC (large).
\textcolor{black}{We compared the BTR with three other rerankers, R2L, BERT, and RoBERTa.}

\textbf{T5GEC}: We used the state-of-the-art model \citep{rothe-etal-2021-simple} as our base model for GEC. This base model inherited the pre-trained T5 version 1.1 model (T5-base) \citep{2020t5} and was fine-tuned as described in Section \ref{seq2seq_gec}. We denote this base model as T5GEC hereafter.
Although the T5 xxl model yielded the most grammatical sentences in \citet{rothe-etal-2021-simple}, it contained 11B parameters and was not suitable for our current experimental environment. Thus, we modeled T5GEC on top of a 248M-parameter T5-base model.
To reproduce the experimental results of \citet{rothe-etal-2021-simple}, we followed their setting and fine-tuned T5GEC once with the cLang-8 dataset. 

\textbf{T5GEC (large)}: To investigate the potential of the BTR for reranking high-quality candidates, we also fine-tuned one larger T5GEC model with a 738M-parameter T5-large structure. We denote this model as T5GEC (large). 

\textbf{R2L}: The decoder of the conventional seq2seq model can generate a target sentence either in a left-to-right or right-to-left direction. Because T5GEC utilized the left-to-right direction, and previous research \citep{sennrich-etal-2016-edinburgh,kiyono-etal-2019-empirical,kaneko-etal-2020-encoder} showed the effectiveness of reranking using the right-to-left model, we followed \citet{kaneko-etal-2020-encoder} to construct four right-to-left T5GEC models, which we denote as R2L. 
R2L reranks candidates based on the sum score of the base model (L2R) and ensembled R2L.

\textcolor{black}{\textbf{BERT}: We followed \citet{kaneko-etal-2019-tmu} to fine-tune four BERT with 334M parameters. During fine-tuning, both source and target sentences 
were annotated with either <0> (ungrammatical) or <1> (grammatical) label for BERT to classify. During inference, the ensembled BERT reranks candidates based on the predicted score for the <1> label.}

\textbf{RoBERTa}: We fine-tuned four 125M parameters RoBERTa to compare our bidirectional Transformer structure with the encoder-only one. During fine-tuning, the source and target sentences were concatenated, and RoBERTa masked and predicted only the target sentence as the BTR. During prediction, the ensembled RoBERTa reranks candidates with the acceptance threshold $\lambda$ as the BTR.

\subsection{Setup for the BTR}
\begin{table*}[t]
\centering
\resizebox{0.9\linewidth}{!}{
\begin{tabular}{lll}
\toprule
\textbf{Model} & \textbf{Inputs} & \textbf{Targets} \\ 
\midrule
\multicolumn{3}{c}{Self-supervised learning for pre-training} \\
\midrule
\textcolor{black}{BERT / RoBERTa} & Thank you \textcolor{blue}{so} <M> me to your party <M> week . & Thank you for inviting me to your party last week .\\
T5 \textcolor{black}{/ R2L} &  Thank you <X> me to your party <Y> week . & <X> for inviting <Y> last <Z> \\
BTR & Thank you <X> me to your party <Y> week . & <X> for <M> \textcolor{blue}{you} last <Z> \\ \midrule
\multicolumn{3}{c}{Supervised learning for fine-tuning} \\
\midrule
BERT & Thank you for inviting me to your party last week . & <1> \\
T5 \textcolor{black}{/ R2L} & Thank you for \textcolor{red}{invite} me to your party last week . & Thank you for inviting me to your party last week . \\ 
\textcolor{black}{BTR / RoBERTa} & Thank you for \textcolor{red}{invite} me to your party last week . & Thank you \textcolor{blue}{so} <M> me to your party <M> week . \\ \bottomrule
\end{tabular}
}
\caption{Examples of data pairs for self-supervised and supervised learning used by each model. The grammatical text is “Thank you for inviting me to your party last week .” <M> denotes a mask token. <X>, <Y>, and <Z> denote sentinel tokens that are assigned unique token IDs. <1> denotes the input sentence is classified as a grammatical sentence. \textcolor{red}{Red} indicates an error in the source sentence while \textcolor{blue}{Blue} indicates a token randomly replaced by the BERT-style masking strategy.}
\label{table:example_data}
\end{table*}

Because there was no pre-trained seq2seq model with a self-attention mechanism for masked language modeling in the decoder, we constructed the BTR using the 248M T5 model (T5-base) and pre-trained it with the Realnewslike corpus \citep{NEURIPS2019_3e9f0fc9}. 
\textcolor{black}{To compare the BTR with R2L, we also constructed R2L using T5-base, and pre-trained both models as follows.}
\textcolor{black}{To speed up pre-training, we initialized the BTR and R2L model parameters with the fine-tuned parameters $\theta$ of T5GEC.}
During pre-training, we followed \citet{2020t5} for self-supervised learning with a span masking strategy. Specifically, 15\% of the tokens in a given sentence were randomly sampled and removed. The input sequence was constructed by the rest tokens while the target sequence was the concatenation of dropped-out tokens. 
\textcolor{black}{An example is provided in Table \ref{table:example_data}.}
We pre-trained the BTR and R2L with $65536=2^{16}$ and 10000 steps, respectively. Because the BTR masked and predicted only 15\% of the tokens in Eq.~(\ref{eq:maxmize_likelyhood}), the true steps for the BTR were $2^{16}\times 0.15 \approx 10000$. We used a maximum sequence length of 512 and a batch size of $2^{20}=1048576$ tokens. In total, we pre-trained $10000 \times 2^{20} \approx 10.5$B tokens, which were less than the pre-trained T5 with 34B tokens. The pre-training for R2L and the BTR took 2 and 13 days, respectively, with 2 NVIDIA A100 80GB GPUs. This indicates the BTR requires more training time and resources than R2L. We \textcolor{black}{provide a} plot \textcolor{black}{of} the pre-training loss in Appendix \ref{sec:appendix_pre-training}.

After pre-training, we successively fine-tuned the BTR with the cLang-8 dataset. Like R2L\textcolor{black}{, BERT,} and RoBERTa, our fine-tuned BTR is the ensemble of four models with random seeds.

\subsection{Datasets}

\begin{table}[t]
\centering
\resizebox{\columnwidth}{!}{
\begin{tabular}{lclr}
\toprule
\textbf{Dataset} & \textbf{Usage}  & \textbf{Lang} & \textbf{\# of data (pairs)} \\ \midrule
Realnewslike & pre-train & EN & 148,566,392 \\
cLang-8 & train & EN & 2,372,119 \\ \midrule
CoNLL-13 (cleaned) & valid & EN& 1,381\\
CoNLL-14 & test &  EN& 1,312 \\
BEA & test &EN & 4,477 \\ 
JFLEG & test & EN &747 \\ \bottomrule
\end{tabular}
}
\caption{Dataset sizes.}
\label{table:data_distribution}
\end{table}

For fair comparison, we pre-trained R2L and the BTR with the Realnewslike corpus. This corpus contains 37 GB of text data and is a subset of the C4 corpus \citep{2020t5}. To shorten the input and target sequences, we split each text into paragraphs. 
During fine-tuning, we followed the steps of \citet{rothe-etal-2021-simple} and regarded the cLang-8 corpus as the training dataset. 

While the CoNLL-13 dataset was used for validation, the standard benchmarks from JFLEG, CoNLL-14, and the BEA test set \citep{bryant-etal-2019-bea} were used for evaluation.
While the CoNLL-14 corpus considers the minimal edit of corrections, JFLEG evaluates the fluency of a sentence. The BEA corpus contains much more diverse English language levels and domains than the CoNLL-14 corpus.
We used a cleaned version of CoNLL-13 with consistent punctuation tokenization styles. Appendix \ref{sec:appendix_cleaning_for_conll_corpus} lists our cleaning steps and the experimental results on the cleaned CoNLL-14 set.
Table \ref{table:data_distribution} summarizes the data statistics. 

\subsection{Evaluation Metrics}
The evaluation on the BEA test set was automatically executed in the official BEA-19 competition in terms of span-based correction $F_{0.5}$ using the ERRANT \citep{bryant-etal-2017-automatic} scorer. For the CoNLL-13 and 14 benchmarks, we evaluated the correction $F_{0.5}$ using the official $M^2$ \citep{dahlmeier-ng-2012-better} scorer. For the JFLEG corpus, we evaluated the GLEU \citep{napoles-etal-2015-ground}.

\textcolor{black}{We report only significant results on the CoNLL-14 set, because the gold data for the BEA test set is unavailable, and the evaluation metric GLEU for the JFLEG test set requires a sampling strategy for multiple references. We used the paired $t$-test to evaluate whether the difference between $\bm{y}_{BTR}$ and $\bm{y}_{base}$ on the CoNLL-14 set is significant, as only limited $\bm{y}_{BTR}$ differed from $\bm{y}_{base}$ among the suggestions from the BTR.}

\subsection{Hyperparameters}
Appendix \ref{sec:appendix_hyperparameters} lists our hyperparameter settings for pre-training and fine-tuning each model.

We followed the setting of \citet{zhang-etal-2021-language} to separately tune $a$ for training and prediction, based on the model performance on the validation dataset with candidates generated by T5GEC. We denote $a$ for training and prediction as $a_{train}$ and $a_{pred}$, respectively. 
\textcolor{black}{The threshold ($\lambda$) for the BTR and RoBERTa was tuned together with $a$.}
\textcolor{black}{We set $a_{train}$ to 20, 0 for the BTR and RoBERTa, respectively, and $a_{pred}$ was set to 5 for all rerankers.}
When $a_{train}=20$, $\lambda$ was set to 0.4 and 0.8 with respect to the candidates generated by T5GEC and T5GEC (large), respectively.
\textcolor{black}{When $a_{train}=0$, $\lambda$ for the RoBERTa was set to 0.1 for the two versions of candidates.}
The experimental results for tuning $a_{train}$, $a_{pred}$, and $\lambda$ are listed in Appendix \ref{sec:appendix_candidates}.

\subsection{Results}

\begin{table*}[t]
\centering
\resizebox{0.9\linewidth}{!}{
\begin{tabular}{lllllllllll}
\toprule
\multirow{2}{*}{\textbf{Model}} & \multicolumn{3}{c}{\textbf{CoNLL-13}}   & \multicolumn{3}{c}{\textbf{CoNLL-14}} & \multicolumn{3}{c}{\textbf{BEA}} & \textbf{JFLEG} \\  \cmidrule(lr){2-4} \cmidrule(lr){5-7} \cmidrule(lr){8-10} \cmidrule(lr){11-11}
 & \multicolumn{1}{c}{\textbf{p}} & \multicolumn{1}{c}{\textbf{r}} & \multicolumn{1}{c}{\textbf{$F_{0.5}$}} & \multicolumn{1}{c}{\textbf{p}} & \multicolumn{1}{c}{\textbf{r}} & \multicolumn{1}{c}{\textbf{$F_{0.5}$}} & \multicolumn{1}{c}{\textbf{p}} & \multicolumn{1}{c}{\textbf{r}} & \multicolumn{1}{c}{\textbf{$F_{0.5}$}} & \textbf{GLEU} \\ \midrule
Oracle & 65.50 & 33.71 & 55.11 & 73.74 & 51.38 & 67.87 & - & - & - & 61.13\\ \hline
T5GEC * & - & - & - & - & - & 65.13  & - & - & 69.38 & - \\
T5GEC & 59.19 & 29.65 & 49.36 & 71.27 & 48.37 & 65.11 & 73.96 & 59.45 & 70.51 & 59.04\\ 
R2L & \textbf{60.94} & 29.14 & 50.02 & \textbf{71.87} & 46.81 & 64.92 & \textbf{75.51} & 58.69 & \bf{71.42} & 58.93 \\
\quad w/o L2R & 59.56 & 28.97 & 49.19 & 71.36 & 46.68 & 64.54 & 73.51 & 57.96 & 69.76 & 58.69 \\
\textcolor{black}{BERT} & 44.53 & \textbf{35.74} & 42.44 & 55.93 & \textbf{53.18} & 55.36  & 49.91 & \textbf{64.37} & 52.26 & 55.69 \\
RoBERTa ($\lambda=0.1$) w/o $a_{train}$ & 59.20 & 29.63 & 49.35 & 71.14 & 48.42 & 65.04 & 74.04 & 59.37 & 70.55 & 59.17 \\
\quad w/o $a_{train}, \lambda$ & 54.83 & 28.88 & 46.48 & 65.64 & 47.24 & 60.90 & 65.85 & 57.71 & 64.05 & 57.49\\
\hline
BTR ($\lambda=0.4$) &59.87 &30.54 & \bf{50.22} & 71.62 & 48.74 & \bf{65.47} & 74.68 & 60.27 & 71.27 & 59.17 \\ 
\quad w/o $\lambda$ & 58.10 & 30.37 & 49.13 & 69.52 & 48.07& 63.82  & 72.69 & 60.71  & 69.93 & \bf{59.52}\\ 
\quad w/o $a_{train}, \lambda$ & 51.30 & 30.94 & 45.34 & 62.83 & 49.03 & 59.48 & 64.35 & 60.74 & 63.60 & 57.62 \\ \bottomrule
\end{tabular}
}
\caption{Results for the models on each dataset with candidates from T5GEC. * indicates the score presented in \citet{rothe-etal-2021-simple}. Bold scores represent the highest (p)recision, (r)ecall, $F_{0.5}$, and GLEU for each dataset.}
\label{table:main_result}
\end{table*}

\begin{table}[t]
\centering
\resizebox{0.8\columnwidth}{!}{
\begin{tabular}{llll}
\toprule
\textbf{Candidate} & \textbf{Accept} & \textbf{Reject} & \textbf{Equal} \\ \midrule
Proportion(\%) & 12.50 & 21.11 & 66.39 \\ \hline
$\bm{y}_{base}$ & 61.67 & \textbf{61.66}$\dagger$ & \textbf{68.78} \\
$\bm{y}_{BTR}$ & \textbf{63.97}$\dagger$  & 57.28 & \textbf{68.78} \\ \bottomrule
\end{tabular}
}
\caption{\textcolor{black}{Results for the BTR ($\lambda=0.4$) on CoNLL-14 with candidates from T5GEC. $\bm{y}_{base}$ and $\bm{y}_{BTR}$ denote the selections by T5GEC and suggestions by the BTR, respectively. $\dagger$ indicates that the difference between $\bm{y}_{BTR}$ and $\bm{y}_{base}$ is significant with a p-value < 0.05. Bold scores represent the highest $F_{0.5}$ for each case.}}
\label{table:pie_scores}
\end{table}

\begin{table}[t]
\resizebox{\columnwidth}{!}{
\begin{tabular}{lllll}
\toprule
\textbf{Model} & \textbf{CoNLL-13} &  \textbf{CoNLL-14} & \textbf{BEA} & \textbf{JFLEG} \\ \midrule
Oracle & 56.44 & 70.08 & - & 63.87 \\ \hline
T5GEC (large) * & - & 66.10 & 72.06 & - \\
T5GEC (large) & 50.79 & 66.83 & 72.15 & 61.88 \\ 
R2L & 50.87 & 66.68 & \bf{72.98} & 61.32 \\ 
\textcolor{black}{RoBERTa ($\lambda=0.1$) w/o $a_{train}$} & 50.76 & \bf{66.85} & 72.20 & 61.85\\ \hline
BTR ($\lambda=0.8$) & \bf{51.00} & 66.57 & 72.41 &  \bf{61.97} \\ \bottomrule
\end{tabular}
}
\caption{Results for the models on each dataset with candidates generated by T5GEC (large). * indicates the score presented in \citet{rothe-etal-2021-simple}. \textcolor{black}{The precision and recall can be found in Appendix \ref{sec:appendix_precision_recall}.}
}
\label{table:T5GEC_large_result}
\end{table}

\pgfplotsset{width=7cm,compat=1.15}
\pgfplotstableread[row sep=\\,col sep=&]{
    Position & T5GEC & R2L & RoBERTa & BTR \\
    1 & 0.594 & 0.996 & 0.229 & 3.361 \\
    2 & 0.698 & 1.407 & 0.157 & 3.445 \\
    3 & 0.413 & 0.931 & 0.228 & 3.339 \\
    4 & 0.515 & 0.885 & 0.385 & 3.361 \\
    5 & 0.572 & 0.947 & 0.250 & 3.329 \\
    6 & 0.395 & 0.764 & 0.181 & 3.287 \\
    7 & 0.459 & 0.935 & 0.084 & 3.358 \\
    8 & 0.180 & 0.763 & 0.351 & 3.391 \\
    9 & 0.762 & 1.180 & 0.133 & 3.401 \\
    10 & 0.449 & 0.729 & 0.130 & 3.398 \\
    11 & 0.573 & 0.893 & 0.216 & 3.323 \\
    12 & 0.356 & 1.040 & 0.229 & 3.373 \\
    13 & 0.449 & 1.021 & 0.094 & 3.388 \\
    14 & 0.623 & 1.184 & 0.092 & 3.380 \\
    15 & 0.679 & 1.099 & 0.051 & 3.437 \\
    16 & 0.324 & 0.821 & 0.074 & 3.385 \\
    17 & 0.138 & 0.314 & 0.008 & 3.383 \\
    18 & 0.140 & 0.142 & 0.000 & 3.409 \\
    19 & 0.116 & 0.117 & 0.000 & 3.656 \\
    20 & 0.069 & 0.080 & 0.000 & 3.350 \\
    }\mydata

\begin{figure}[t]
\begin{adjustbox}{width=1\columnwidth}
\begin{tikzpicture}
    \pgfplotsset{set layers}
    \begin{axis}[
            scale only axis,
            legend pos=outer north east,
            x post scale=2,
            y post scale=1,
            ylabel=Cross-entropy loss,
            xlabel=Token position ($j$),
            ymin=0, ymax=3.7,
            xmin=1, xmax=20,
            symbolic x coords={1,2,3,4,5,6,7,8,9,10,11,12,13,14,15,16,17,18,19,20},
        ]
        \addlegendimage{/pgfplots/refstyle=plot_green}\addlegendentry{BTR}
        \addlegendimage{/pgfplots/refstyle=plot_olive}\addlegendentry{R2L}
        \addlegendimage{/pgfplots/refstyle=plot_bleu}\addlegendentry{T5GEC}
        \addlegendimage{/pgfplots/refstyle=plot_red}\addlegendentry{RoBERTa}
        \addplot+[sharp plot, mark=otimes, blue] table[x=Position,y=T5GEC]{\mydata};\label{plot_bleu}
        \addplot+[sharp plot, mark=triangle, green] table[x=Position,y=BTR]{\mydata};\label{plot_green}
        \addplot+[sharp plot, mark=square, olive] table[x=Position,y=R2L]{\mydata};\label{plot_olive}
        \addplot+[sharp plot, mark=diamond, red] table[x=Position,y=RoBERTa]{\mydata};\label{plot_red}
    \end{axis}
\end{tikzpicture}
\end{adjustbox}
\caption{Cross-entropy loss of $y_j$ versus $j$. The loss was averaged over CoNLL-14's 149 tokenized utterances with length in interval $[18, 20]$ (including <eos>). } 
\label{fig:cross_entropy}
\end{figure}
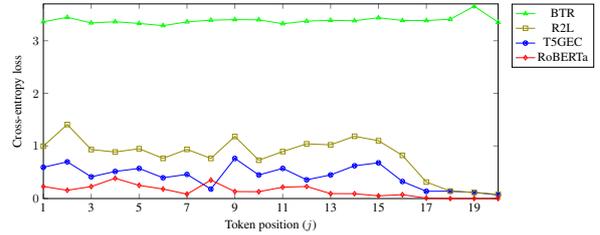

\pgfplotsset{width=7cm,compat=1.15}
\pgfplotstableread[row sep=\\,col sep=&]{
    Position & BTRatrain0 & BTRatrain5 & BTRatrain10 & BTRatrain20 & R2L & BERT & RoBERTaatrain0 & RoBERTaatrain5 & RoBERTaatrain10 & RoBERTaatrain20 \\
    1 & 0.2399 & 0.6176 & 0.6629 & 0.6876 & 0.2673 & 0.255 & 0.2235 & 0.2262 & 0.2250 & 0.2241\\
    2 & 0.2139 & 0.1898 & 0.1856 & 0.177 & 0.2155 & 0.2299 & 0.2161 & 0.2172 & 0.2168 & 0.2162\\
    3 & 0.1975 & 0.0959 & 0.0812 & 0.0745 & 0.1887 & 0.2084 & 0.2016 & 0.2011 & 0.2014 & 0.2015\\
    4 & 0.1830 & 0.0600 & 0.0455 & 0.0401 & 0.1734 & 0.1775 & 0.1879 & 0.1865 & 0.1872 & 0.1876\\
    5 & 0.1656 & 0.0366 & 0.0249 & 0.0207 & 0.1552 & 0.1292 & 0.1709 & 0.169 & 0.1696 & 0.1705\\
    }\mydatarankdistribution

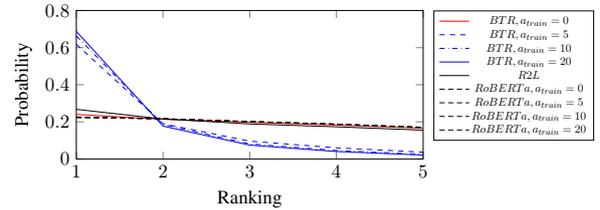
\begin{figure}[t]
    \begin{adjustbox}{width=1\columnwidth}
        \begin{tikzpicture}
            \begin{axis}[
                scale only axis,
                legend pos=outer north east,
                legend style={nodes={scale=0.6, transform shape}},
                x post scale=1,
                y post scale=0.5,
                ylabel=Probability,
                xlabel=Ranking,
                ymin=0, ymax=0.8,
                xmin=1, xmax=5,
            ]
            \addlegendimage{/pgfplots/refstyle=plot_BTRatrain0_ranking}\addlegendentry{$BTR, a_{train}=0$}
            \addlegendimage{/pgfplots/refstyle=plot_BTRatrain5_ranking}\addlegendentry{$BTR, a_{train}=5$}
            \addlegendimage{/pgfplots/refstyle=plot_BTRatrain10_ranking}\addlegendentry{$BTR, a_{train}=10$}
            \addlegendimage{/pgfplots/refstyle=plot_BTRatrain20_ranking}\addlegendentry{$BTR, a_{train}=20$}
            \addlegendimage{/pgfplots/refstyle=plot_R2L_ranking}\addlegendentry{$R2L$}
            \addlegendimage{/pgfplots/refstyle=plot_RoBERTaatrain0_ranking}\addlegendentry{$RoBERTa, a_{train}=0$}
            \addlegendimage{/pgfplots/refstyle=plot_RoBERTaatrain5_ranking}\addlegendentry{$RoBERTa, a_{train}=5$}
            \addlegendimage{/pgfplots/refstyle=plot_RoBERTaatrain10_ranking}\addlegendentry{$RoBERTa, a_{train}=10$}
            \addlegendimage{/pgfplots/refstyle=plot_RoBERTaatrain20_ranking}\addlegendentry{$RoBERTa, a_{train}=20$}
            \addplot+[sharp plot, mark=none, red] table[x=Position,y=BTRatrain0]{\mydatarankdistribution};\label{plot_BTRatrain0_ranking}
            \addplot+[sharp plot, mark=none, blue, dashed] table[x=Position,y=BTRatrain5]{\mydatarankdistribution};\label{plot_BTRatrain5_ranking}
            \addplot+[sharp plot, mark=none, blue, dash dot] table[x=Position,y=BTRatrain10]{\mydatarankdistribution};\label{plot_BTRatrain10_ranking}
            \addplot+[sharp plot, mark=none, blue] table[x=Position,y=BTRatrain20]{\mydatarankdistribution};\label{plot_BTRatrain20_ranking}
            \addplot+[sharp plot, mark=none, black] table[x=Position,y=R2L]{\mydatarankdistribution};\label{plot_R2L_ranking}
            \addplot+[sharp plot, mark=none, black] table[x=Position,y=RoBERTaatrain0]{\mydatarankdistribution};\label{plot_RoBERTaatrain0_ranking}
            \addplot+[sharp plot, mark=none, black] table[x=Position,y=RoBERTaatrain5]{\mydatarankdistribution};\label{plot_RoBERTaatrain5_ranking}
            \addplot+[sharp plot, mark=none, black] table[x=Position,y=RoBERTaatrain10]{\mydatarankdistribution};\label{plot_RoBERTaatrain10_ranking}
            \addplot+[sharp plot, mark=none, black] table[x=Position,y=RoBERTaatrain20]{\mydatarankdistribution};\label{plot_RoBERTaatrain20_ranking}
        \end{axis}
        \end{tikzpicture}
    \end{adjustbox}
\caption{\textcolor{black}{Average probability for each rank on the CoNLL-14 test set. The top-5 candidate sentences were generated by T5GEC.}}
\label{fig:probability_distribution_of_ranking}
\end{figure}

\textcolor{black}{Table \ref{table:main_result} presents our main results.\footnote{The mean and standard deviation results of the BTR, R2L, and RoBERTa are listed in Appendix \ref{section:appendix_mean_std}.}}
While reranking by R2L yielded the highest $F_{0.5}$ score of 71.42 on the BEA test set, it yielded only a lower score than the BTR ($\lambda=0.4$) on CoNLL-14 and JFLEG test sets.
Meanwhile, the improvements brought by R2L depended on the beam searching score from L2R, suggesting that the unidirectional representation offers fewer gains compared to the bidirectional representation from the BTR.
\textcolor{black}{Reranking candidates by BERT resulted in the lower $F_{0.5}$ and GLEU scores than T5GEC. This may be because BERT considers only the target sentence and ignores the relationship between the source and the target.}
The BTR ($\lambda=0.4$) achieved an $F_{0.5}$ score of 71.27 on the BEA test set.\footnote{Experimental results in more details for different CEFR levels and error types can be found in Appendix \ref{sec:appendix_cefr_results}.}
On the CoNLL-14 test set, the BTR ($\lambda=0.4$) attained the highest $F_{0.5}$ score of 65.47, with improvements of 0.36 points from T5GEC.
The use of the threshold and negative candidates played an important role in the BTR. 
Without these two mechanisms, the BTR achieved only 59.48 and 63.60 $F_{0.5}$ scores, respectively, on the CoNLL-14 and BEA test sets, which were lower than those of the original selection.
In the meantime, the BTR without the threshold could achieve the highest GLEU score of 59.52 on the JFLEG corpus, which indicates $\lambda=0.4$ is too high for the JFLEG corpus.
This is because of the different distributions and evaluation metrics between the CoNLL-13 and JFLEG corpus, as proved in Appendix \ref{sec:appendix_relation_between_lambda_pre_rec}.
Compared to RoBERTa ($\lambda=0.1$) w/o $a_{train}$ of the encoder-only structure, the BTR ($\lambda=0.4$) can achieve higher $F_{0.5}$ scores on CoNLL-13, 14, and BEA test sets, and a competitive GLEU score on the JFLEG corpus. These results show the benefit of using the Transformer with the encoder-decoder architecture in the BTR.

\textcolor{black}{Table \ref{table:pie_scores} demonstrates the effect of using $\lambda$.}
\emph{Equal} denotes the suggestion $\bm{y}_{BTR}$ is exactly $\bm{y}_{base}$. \emph{Accept} denotes $\bm{y}_{BTR}$ satisfies Eq.~(\ref{eq:accpet_threshold}) and $\bm{y}_{BTR}$ will be the final selection, while \emph{Reject} denotes $\bm{y}_{BTR}$ does not satisfy the equation and $\bm{y}_{base}$ is still the final selection. 
Most of the final selections belonged to \emph{Equal} and achieved the highest $F_{0.5}$ score of 68.78.
This indicates the sentences in \emph{Equal} can be corrected easily by both the BTR ($\lambda=0.4$) and T5GEC. 
Around $1/3$ of the new suggestions proposed by the BTR ($\lambda=0.4$) were accepted and achieved an $F_{0.5}$ score of 63.97, which was a 2.3-point improvement from $\bm{y}_{base}$.
However, around $2/3$ of the new suggestions were not accepted, and the original selection by T5GEC resulted in a higher $F_{0.5}$ score than these rejected suggestions. 
These results show that, among the new suggestions, the BTR was confident only for some suggestions. The confident suggestions tended to be more grammatical, whereas the unconfident suggestions tended to be less grammatical than the original selections. Appendix {\ref{sec:appendix_relation_between_lambda_pre_rec} shows the analysis.}

\textcolor{black}{Table \ref{table:T5GEC_large_result} lists the performances when reranking high-quality candidates.}
\textcolor{black}{While R2L still achieved the highest $F_{0.5}$ score on the BEA test set, it was less effective than the BTR on the JFLEG corpus.}
\textcolor{black}{Although the BTR ($\lambda=0.8$) used only 248M parameters and was trained with the candidates generated by T5GEC, it could rerank candidates from T5GEC (large) and achieve 61.97 GLEU and 72.41 $F_{0.5}$ scores on the JFLEG and BEA test sets, respectively.}
This finding indicates the sizes of the BTR and the base model do not need to be consistent, and a smaller BTR can also work as a reranker for a larger base model. 
\textcolor{black}{RoBERTa ($\lambda=0.1$) w/o $a_{train}$ achieved the highest $F_{0.5}$ score of 66.85 on the CoNLL-14 corpus with only 0.02-point improvement from T5GEC (large), which reflects the difficulty in correcting uncleaned sentences.}

\textcolor{black}{To investigate the difference among R2L, RoBERTa ($\lambda=0.1$) w/o $a_{train}$, and the BTR ($\lambda=0.4$), we compared the precision and recall of the three rerankers in Table \ref{table:main_result}.}
In most cases, R2L tended to improve the precision but lower the recall from T5GEC.
\textcolor{black}{The improvements brought by RoBERTa from T5GEC for both precision and recall are limited.}
Meanwhile, the BTR could improve both precision and recall from the original ranking. Because T5GEC already achieved a relatively high precision and low recall, there was more room to improve recall, which was demonstrated by the BTR. Figure \ref{fig:cross_entropy} shows both T5GEC and R2L have a relatively high cross-entropy loss for tokens at the beginning positions and a low loss for tokens at the ending positions, even though the loss of R2L was the sum of two opposite decoding directions. This may be because the learning by the auto-regressive models for the latest token was over-fitting and for the global context was under-fitting, as \citet{qi-etal-2020-prophetnet} indicated.
\textcolor{black}{RoBERTa has a flatter loss with less sharp positions than T5GEC and R2L.}
Meanwhile, the BTR has a flat loss, which is ideal for reranking candidate sentences with length normalization, as suggested by \citet{salazar-etal-2020-masked}.
\textcolor{black}{Figure \ref{fig:probability_distribution_of_ranking} shows the probability distribution of reranking. When $a_{train} > 0$, the probability distribution of the BTR becomes peaked, which indicates that using Eq.~(\ref{eq:maxmize_likelyhood}) to minimize the unlikelihood could increase the probability gap between the 1st-ranked candidate and the rest. Compared with the BTR, when $a_{train} > 0$, the probability distribution of RoBERTa is as flat as $a_{train} = 0$, which suggests the effectiveness of the encoder-decoder structure compared with the encoder-only one when minimizing unlikelihood.}

\section{Conclusion}
We proposed a bidirectional Transformer reranker (BTR) to rerank several top candidates generated by a pre-trained seq2seq model for GEC. For a fully pre-trained model, T5-base, the BTR could achieve 65.47 and 71.27 $F_{0.5}$ scores on the CoNLL-14 and BEA test sets. Our experimental results showed that the BTR on top of T5-base with limited pre-training steps could improve both precision and recall for candidates from T5-base.
\textcolor{black}{Since using negative sampling for the BTR generates a peaked probability distribution for ranking, introducing a threshold $\lambda$ benefits the acceptance of the suggestion from the BTR.}
Furthermore, the BTR on top of T5-base could rerank candidates generated from T5-large and yielded better performance. This finding suggests the effectiveness of the BTR even in experiments with limited GPU resources. While the BTR in our experiments lacked sufficient pre-training, it should further improve the performance with full pre-training for reranking in future.

\section{Limitations}
As mentioned in the previous section, up until now, there has not been a fully pre-trained seq2seq model with a BERT-style self-attention mechanism in the decoder, while the vanilla seq2seq model tends to use a left-to-right or right-to-left unidirectional self-attention.
Therefore, utilizing our proposed Bidirectional Transformer Reranker (BTR) to rerank candidates from a pre-trained vanilla seq2seq model requires additional pre-training steps, which cost both time and GPU resources. 
Because the BTR masks and predicts only 15\% of the tokens in Eq.~(\ref{eq:maxmize_likelyhood}), it requires more training steps than a vanilla seq2seq model. In addition, during fine-tuning, the BTR also requires additional $a_{train}$ negative samples, which makes the fine-tuning longer. Furthermore, tuning $a_{train}$ will be inefficient if the training is slow. In other words, training an effective BTR requires much more time than training a vanilla seq2seq model.

As a reranker, the performance of the BTR depends on the quality of candidates. There is no room for improvement by the BTR if no candidate is more grammatical than the original selection.

\bibliography{anthology, custom}
\bibliographystyle{acl_natbib}
\input{appendix.tex}
\end{document}

%% file: appendix.tex
\appendix
\section{Computation for $\tilde{s}_{j}$ in Transformer}
\label{sec:appendix_computation_for_sj}
Let $\mathrm{FNN}$ denote a feed-forward layer and $\mathrm{Attn}(q, K, V )$ the attention layer, where $q$, $K$, and $V$ indicate the query, key, and value, respectively. We assume the decoder consists of $L$ layers. To compute $\tilde{s}_{j}$, the encoder first encodes $\bm{x}$ into its representation $\widetilde{H}$. Then, for $\ell \in L$, the hidden state $\tilde{s}_{j}^{\ell}$ of the $\ell$-th layer in the decoder is computed by 
\begin{align}
  s_{j}^{\ell} &= \mathrm{Attn}_{s}(\tilde{s}_{j}^{\ell-1}, \widetilde{S}_{\leq{j}}^{\ell-1}, \widetilde{S}_{\leq{j}}^{\ell-1}), \label{eq:Attns_seq2seq} \\ 
    \hat{s}_{j}^{\ell} &= \mathrm{Attn}_{c}(s_{j}^{\ell}, \widetilde{H}, \widetilde{H}), \label{eq:attnc} \\ 
    \tilde{s}_{j}^{\ell} &= \mathrm{FNN}(\hat{s}_{j}^{\ell}), \label{eq:FFN}
\end{align}
where $\tilde{s}_{j}^{0}$ is the embedding of the token $y_{j-1}$ and $\tilde{s}_{1}$ is the state for the special token <s>, that indicates the start of a sequence. $\widetilde{S}_{\leq{j}}^{\ell-1}$ denotes a set of hidden states $(\tilde{s}_{1}^{\ell-1},\ldots, \tilde{s}_{j}^{\ell-1})$. $\mathrm{Attn}_{s}$ and $\mathrm{Attn}_{c}$ indicate the self-attention and cross-attention layers, respectively. A causal attention mask can be used to compute $S^{\ell}$ in parallel, as in Figure \ref{fig:causal_attention}.

\section{Computation for 
\textcolor{black}{$\tilde{h}_{k}^{\ell}$}
in BERT}
\label{sec:appendix_computation_for_hkl}
Assuming the model consists of $L$ layers. Without the cross-attention, 
\textcolor{black}{$\tilde{h}_{k}^{\ell}$}
is the feed-forward result of 
\textcolor{black}{$h_{k}^{\ell}$: }
\begin{eqnarray}
  h_{k}^{\ell} &= \mathrm{Attn}_{s}(\tilde{h}_{k}^{\ell-1}, \widetilde{H}_{\backslash \kappa }^{\ell - 1}, \widetilde{H}_{\backslash \kappa }^{\ell - 1}), \label{eq:Attns_bert}
\end{eqnarray}
where $\tilde{h}_{k}^{0}$ is the embedding of the $k$-th token in $\bm{x}_{\backslash \kappa}$ and $\widetilde{H}_{\backslash \kappa }^{\ell-1}=(\tilde{h}_{1}^{\ell-1}, \ldots,\tilde{h}_{m}^{\ell-1})$ denotes a set of hidden states for $\bm{x}_{\backslash \kappa}$. Compared with $s_{j}^{\ell}$, $h_{k}^{\ell}$ utilizes both the left and right sides of the context of the masked token $x_{k}$ to capture deeper representations.

\section{Procedure for Prediction}
\label{sec:appendix_procedure}
\begin{figure}[ht]
    \includegraphics[width=1\columnwidth]{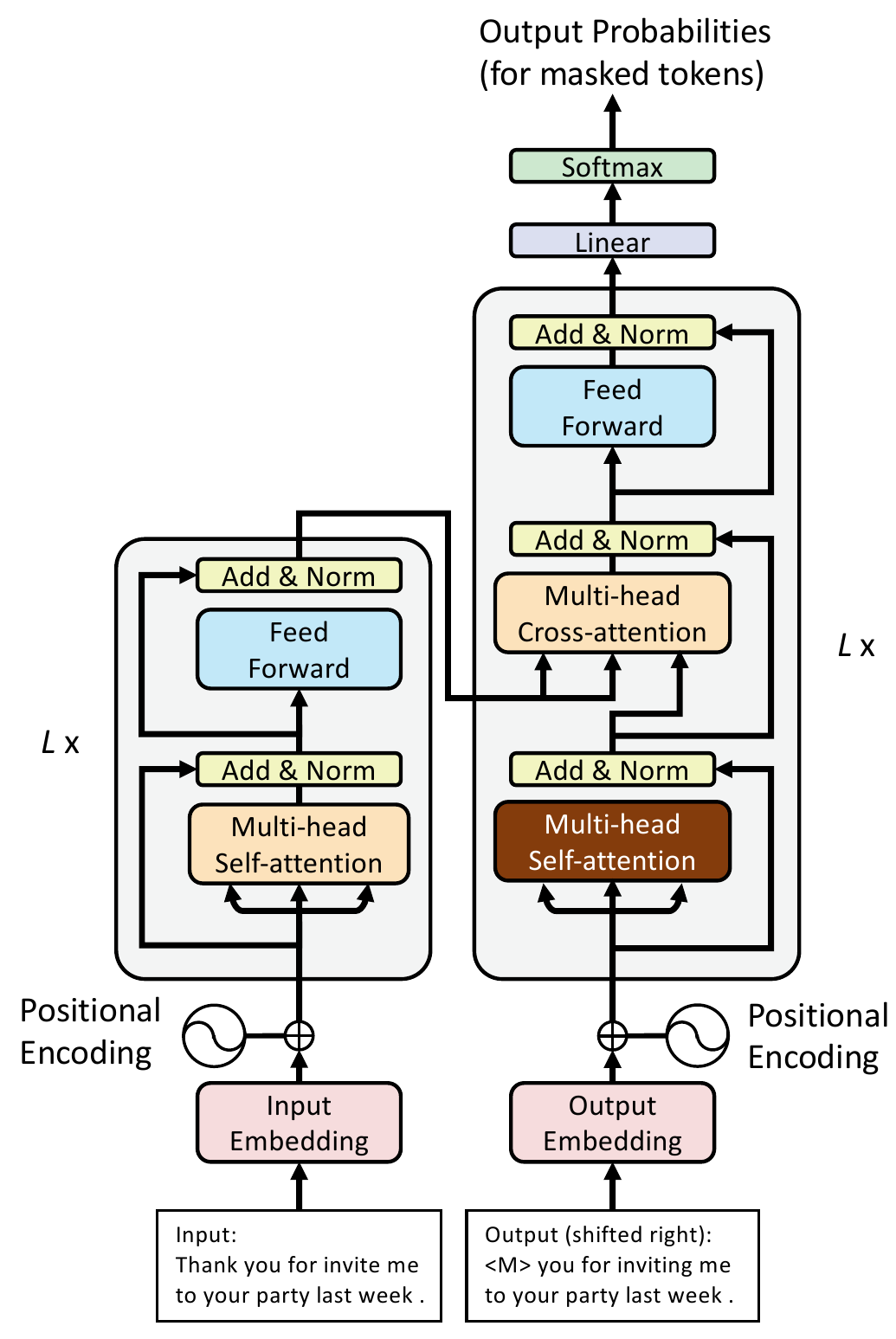}
    \caption{Bidirectional Transformer architecture. The left and right columns indicate the encoder and decoder, respectively. The self-attention mechanism in the decoder utilizes the fully-visible mask (Figure \ref{fig:fully_attention}), unlike the conventional Transformer \cite{NIPS2017_3f5ee243}.}
    \label{fig:bi_transformer}
\end{figure}

Figure \ref{fig:bi_transformer} shows our procedure for prediction. 

\section{Pre-training Loss}
\label{sec:appendix_pre-training}
\begin{figure}[ht]
    \begin{subfigure}[t]{0.45\columnwidth}
        \centering
        \begin{adjustbox}{width=1\columnwidth}
        \begin{tikzpicture}
        \begin{axis}[
        ymin=1.2, ymax=2.5,
        xmin=0, xmax=9900,
        xlabel=Step,
        ylabel=Loss,
        x post scale=1,
        y post scale=0.8,
        ]
    	\addplot[mark = none, sharp plot, orange]
    	table{
    	X Y
        101 4.694
        201 3.513
        301 3.002
        401 2.709
        501 2.514
        601 2.374
        701 2.272
        801 2.194
        901 2.131
        1001 2.078
        1101 2.033
        1201 1.994
        1301 1.960
        1401 1.930
        1501 1.903
        1601 1.879
        1701 1.858
        1801 1.838
        1901 1.820
        2001 1.803
        2101 1.788
        2201 1.774
        2301 1.760
        2401 1.748
        2501 1.736
        2601 1.725
        2701 1.714
        2801 1.704
        2901 1.695
        3001 1.686
        3101 1.677
        3201 1.669
        3301 1.661
        3401 1.654
        3501 1.647
        3601 1.640
        3701 1.633
        3801 1.627
        3901 1.621
        4001 1.615
        4101 1.610
        4201 1.605
        4301 1.599
        4401 1.594
        4501 1.590
        4601 1.585
        4701 1.581
        4801 1.576
        4901 1.572
        5001 1.568
        5101 1.564
        5201 1.560
        5301 1.556
        5401 1.553
        5501 1.549
        5601 1.545
        5701 1.542
        5801 1.539
        5901 1.535
        6001 1.532
        6101 1.529
        6201 1.526
        6301 1.523
        6401 1.520
        6501 1.518
        6601 1.515
        6701 1.512
        6801 1.509
        6901 1.507
        7001 1.504
        7101 1.502
        7201 1.499
        7301 1.497
        7401 1.495
        7501 1.492
        7601 1.490
        7701 1.488
        7801 1.485
        7901 1.483
        8058 1.311
        8158 1.310
        8258 1.309
        8358 1.309
        8458 1.308
        8558 1.307
        8658 1.307
        8758 1.307
        8858 1.306
        8958 1.306
        9058 1.305
        9158 1.304
        9258 1.304
        9358 1.303
        9458 1.303
        9558 1.302
        9658 1.302
        9758 1.301
        9858 1.301
        9958 1.300
    	};
    	\addlegendentry{R2L}
    	\end{axis}
        \end{tikzpicture}
        \end{adjustbox}
    \end{subfigure}
    \begin{subfigure}[t]{0.45\columnwidth}
        \centering
        \begin{adjustbox}{width=1\columnwidth}
        \begin{tikzpicture}
        \begin{axis}[
        ymin=0.8, ymax=1.1,
        xmin=0, xmax=65000,
        xlabel=Step,
        ylabel=Loss,
        x post scale=1,
        y post scale=0.8,
        ]
    	\addplot[mark = none, sharp plot, blue]
    	table{
    	X Y
    	700 2.046
    	1400 1.384
    	2100 1.264
    	2800 1.194
    	3500 1.155
    	4200 1.125
    	4900 1.102
    	5600 1.087
    	6300 1.072
    	7000 1.058
    	7700 1.047
    	8400 1.037
    	9100 1.025
    	9800 1.017
    	10500 1.011
    	11200 1.006
    	11900 0.996
    	12600 0.991
    	13300 0.987
    	14000 0.982
    	14700 0.979
    	15400 0.973
    	16100 0.969
    	16800 0.965
    	17500 0.956
    	18200 0.953
    	18900 0.949
    	19600 0.947
    	20300 0.945
    	21000 0.943
    	21700 0.942
    	22400 0.94
    	23100 0.938
    	23800 0.936
    	24500 0.935
    	25200 0.931
    	25900 0.929
    	26600 0.927
    	27300 0.926
    	28000 0.924
    	28700 0.923
    	29400 0.922
    	30100 0.92
    	30800 0.919
    	31500 0.917
    	32200 0.916
    	32900 0.913
    	33600 0.912
    	34300 0.91
    	35000 0.909
    	35700 0.908
    	36400 0.907
    	37100 0.906
    	37800 0.904
    	38500 0.903
    	39200 0.901
    	39900 0.9
    	40600 0.901
    	41300 0.896
    	42000 0.895
    	42700 0.893
    	43400 0.892
    	44100 0.891
    	44800 0.889
    	45500 0.889
    	46200 0.888
    	46900 0.887
    	47600 0.886
    	48300 0.885
    	49000 0.883
    	49700 0.883
    	50400 0.883
    	51100 0.881
    	51800 0.881
    	52500 0.88
    	53200 0.879
    	53900 0.878
    	54600 0.878
    	55300 0.877
    	56000 0.876
    	56700 0.876
    	57400 0.869
    	58100 0.867
    	58800 0.866
    	59500 0.865
    	60200 0.864
    	60900 0.863
    	61600 0.863
    	62300 0.862
    	63000 0.862
    	63700 0.861
    	64400 0.861
    	65100 0.861
    	};
    	\addlegendentry{BTR}
    	\end{axis}
        \end{tikzpicture}
        \end{adjustbox}
    \end{subfigure}
\caption{Pre-training loss for R2L (left) and the BTR (right).} 
\label{fig:pre-training_loss}
\end{figure}
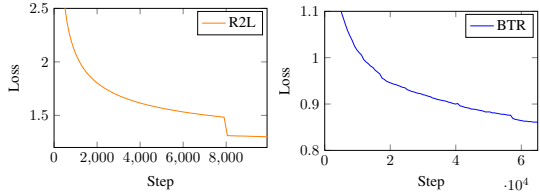

Figure \ref{fig:pre-training_loss} shows the pre-training loss for R2L and the BTR on the Realnewslike corpus. The training loss of R2L suddenly dropped from 1.48 to 1.3 after the first epoch (7957 steps).

\section{Cleaning for CoNLL Corpus}
\label{sec:appendix_cleaning_for_conll_corpus}

\begin{table}[ht]
\centering
\resizebox{1\columnwidth}{!}{
\begin{tabular}{llll}
\toprule
\textbf{Model} & \textbf{Precision} & \textbf{Recall} & \textbf{$F_{0.5}$} \\ \midrule
Oracle & 80.62 & 51.98 & 72.62 \\ \midrule
T5GEC  & 78.01 & 48.57 & 69.58 \\ 
R2L  & \textbf{78.81} & 46.83 & 69.34 \\
\quad w/o L2R  & 77.69 & 46.55 & 68.52 \\
BERT  & 58.84 & \textbf{53.53} & 57.70 \\ 
RoBERTa ($\lambda=0.1$) w/o $a_{train}$ & 77.86 & 48.62 & 69.50 \\
\quad w/o $a_{train}, \lambda$ & 71.07 & 47.36 & 64.60 \\
\midrule
BTR ($\lambda=0.4$)  & 78.52 & 48.82 & \bf{70.00} \\ 
\quad w/o $\lambda$ & 76.02 & 48.30 & 68.19 \\ 
\quad w/o $a_{train}, \lambda$ & 67.44 & 49.45 & 62.87\\\bottomrule
\end{tabular}
}
\caption{Results for the models on the cleaned CoNLL-14 corpus with candidates from T5GEC. Bold scores represent the highest precision, recall, and $F_{0.5}$.}
\label{table:main_result_conll14}
\end{table}

\begin{table}[ht]
\centering
\resizebox{0.9\columnwidth}{!}{
\begin{tabular}{lc}
\toprule
\textbf{Model} & \textbf{$F_{0.5}$} \\ \midrule
R2L & 69.36 $\pm$ 0.13 \\
RoBERTa ($\lambda=0.1$) w/o $a_{train}$  & 68.12 $\pm$ 2.39 \\
BTR ($\lambda=0.4$) & \textbf{69.80} $\pm$  0.18  \\ \bottomrule
\end{tabular}
}
\caption{The mean $\pm$ std results on the cleaned CoNLL-14 corpus with candidates from T5GEC. Bold scores represents the highest mean.}
\label{table:mean_std_conll14}
\end{table}

\begin{table}[ht]
\centering
\resizebox{1\columnwidth}{!}{
\begin{tabular}{llll}
\toprule
\textbf{Model} & \textbf{Precision} & \textbf{Recall} & \textbf{$F_{0.5}$} \\ \midrule
Oracle & 82.01 & 54.19 & 74.38 \\ \midrule
T5GEC (large) & 79.27 & 49.91 &70.92 \\ 
R2L & \textbf{79.72} & 48.71 &70.72 \\ 
\textcolor{black}{RoBERTa ($\lambda=0.1$) w/o $a_{train}$} & 79.30 & 49.91 & 70.94\\ \midrule
BTR ($\lambda=0.8$) & 79.65 & \textbf{49.98} &\bf{71.20} \\ \bottomrule
\end{tabular}
}
\caption{Results for the models on the cleaned CoNLL-14 corpus with candidates from T5GEC (large). Bold scores represent the highest precision, recall, and $F_{0.5}$.}
\label{table:T5GEC_large_result_conll14}
\end{table}

The original texts of CoNLL-13 and 14 contain several styles of punctuation tokenization, such as ``DementiaToday,2012'' and ``known , a''. While these punctuation styles with/without spaces are not considered grammatical errors by a human, they are often identified as errors by automatic GEC scorers.
Moreover, while most of the sequences in CoNLL-14 are of sentence-level, several sequences are of paragraph-level due to the punctuation without spaces. 
In this research, we cleaned the texts of CoNLL-13 and 14 using the ``en\_core\_web\_sm'' tool in spaCy \citep{honnibal2020spacy} so that all punctuation included spaces. The paragraph-level sequences were split into sentences with respect to the position of full stops. The cleaned CoNLL-14 corpus contains 1326 pairs of data.

Tables \ref{table:main_result_conll14}, \ref{table:mean_std_conll14} and \ref{table:T5GEC_large_result_conll14} show the experimental results on the cleaned CoNLL-14 corpus.

\section{Hyperparameters}
\label{sec:appendix_hyperparameters}
\begin{table*}[ht]
\centering
\resizebox{1\linewidth}{!}{
\begin{tabular}{lccccccc}
\toprule
\textbf{Hyperparameters} & \textbf{T5GEC} & \textbf{BERT} & \textbf{RoBERTa} & \textbf{R2L (pretrain)} & \textbf{R2L (finetune)} & \textbf{BTR (pretrain)} & \textbf{BTR (finetune)} \\ \midrule
\# of updates & 15 (epochs) & 15 (epochs) & 15 (epochs) & 10000 & 15 (epochs) & 65536 & 15 (epochs) \\
Max src / tgt length (train)& 128 & 128  & 128 & 512 & 128 & 512 &  128 \\
Max src / tgt length (eval) & 512 & 1 & 512 & 512 & 512 &512 & 512 \\
$a_{train}$ & - & - & \{0, 5, 10, 20\} & - & - & - & \{0, 5, 10, 20\} \\
$a_{pred}$ & - & \{5, 10, 15, 20\} & \{5, 10, 15, 20\} & - & \{5, 10, 15, 20\} & - & \{5, 10, 15, 20\} \\
Threshold ($\lambda$)  & - & - & \{0, 0.1, 0.2,\ldots,0.9\} & - & - & - & \{0, 0.1, 0.2,\ldots,0.9\}\\
\bottomrule
\end{tabular}
}
\caption{\textcolor{black}{Used hyperparameters.}}
\label{table:hyperparameters}
\end{table*}

\begin{table*}[ht]
\centering
\resizebox{0.8\linewidth}{!}{
\begin{tabular}{ll}
\toprule
\textbf{Used artifacts} & \textbf{Note}                                                                      \\ \midrule
T5-base        & \url{https://huggingface.co/google/t5-v1\_1-base}        \\
T5-large       & \url{https://huggingface.co/google/t5-v1\_1-large}       \\
T5GEC          &\url{https://github.com/google-research-datasets/clang8/issues/3} \\
RoBERTa        & \url{https://huggingface.co/roberta-base}                \\
BERT           & \url{https://huggingface.co/bert-large-cased}  \\
cLang-8        & \url{https://github.com/google-research-datasets/clang8} \\
CoNLL-13       & File \emph{revised/data/official-preprocessed.m2}        \\
CoNLL-14       & File \emph{alt/official-2014.combined-withalt.m2}        \\
JFLEG          & File \emph{test/test.src}                                \\
ERRANT         & \url{https://github.com/chrisjbryant/errant}     \\
Fairseq        & \url{https://github.com/facebookresearch/fairseq/} \\
HuggingFace & \url{https://github.com/huggingface/transformers/} \\
BEA-19 competition & \url{https://competitions.codalab.org/competitions/20229} \\
\bottomrule
\end{tabular}
}
\caption{Used artifacts.}
\label{table:used_artifacts}
\end{table*}

Table \ref{table:hyperparameters} lists the hyperparameter settings used for each model. And Table \ref{table:used_artifacts} lists the used artifacts.
The setting for T5GEC (large) was the same as T5GEC.
We followed the setting of \citet{kaneko-etal-2019-tmu} to use a 0.0005 learning rate for the BERT reranker. We used a 0.0001 learning rate for the RoBERTa reranker. For both BERT and RoBERTa, we utilized the adam optimizer, ``inverse square root'' learning rate schedule, and 1.2 epochs warm-up steps.
For other models based on a T5 structure, we used a 0.001 learning rate and adafactor optimizer. The batch size was 1048576 tokens for all models.
We used the Fairseq \citep{ott-etal-2019-fairseq} and HuggingFace \citep{wolf-etal-2020-transformers} to reproduce all models and run the BTR.

\section{Candidate and Threshold Tuning}
\label{sec:appendix_candidates}
Following \citet{zhang-etal-2021-language}, we tuned $a$ for training and predicting separately on the validation dataset with candidates generated by T5GEC. Table \ref{table:size_of_training_candidates} lists the size of training data with candidates generated by T5GEC.
When tuning $a_{train} \in \{0, 5, 10, 20\}$\footnote{Setting $a$ to 0 indicates training with only gold data.} for the BTR, $a_{pred}$ was fixed to 5. 
Because the BTR with $\lambda=0.4$ and $a_{train}=20$ achieved the highest score as shown in Table \ref{table:train_topa_result}, $a_{train}$ was fixed to 20, this BTR was also used to tune $a_{pred} \in \{5, 10, 15, 20\}$. 
When tuning $a_{train} \in \{0, 5, 10, 20\}$ for RoBERTa, $a_{pred}$ was fixed to 5. The results in Tables \ref{table:train_topa_result} and \ref{table:train_topa_result_RoBERTa} indicate the different distributions of $F_{0.5}$ score between RoBERTa and the BTR.
To investigate the reason, we compared the training loss and $F_{0.5}$ score of RoBERTa with the BTR. Figure \ref{fig:diff_atrain} shows the comparison. Different from the BTR, when using negative sampling ($a_{train} > 0$) for training RoBERTa, the $F_{0.5}$ score on the CoNLL-13 corpus decreased with the epoch increasing. The training loss of RoBERTa also dropped suddenly after finishing the first epoch. This result suggests that negative sampling in the GEC task for an encoder-only structure leads in the wrong direction in learning representations from the concatenated source and target.
And therefore, we fixed $a_{train}$ to 0 for RoBERTa. This RoBERTa was also used to tune $a_{pred} \in \{5, 10, 15, 20\}$.
The results in Tables \ref{table:eval_topa_result_BTR}, \ref{table:eval_topa_result_RoBERTa}, and \ref{table:eval_topa_result_R2L} show that when $a_{pred}$ was set to 5, the BTR, R2L, RoBERTa, and BERT attained their highest scores on the CoNLL-13 corpus. Thus, $a_{pred}$ was fixed to 5 in our experiments.

\begin{table}[t]
\centering
\begin{tabular}{lr}
\toprule
$a_{train}$ & \textbf{\# of training data (pairs)} \\ \midrule
0 & 2,371,961 \\ 
5 & 13,727,133 \\ 
10 & 22,396,187\\
20 & 30,423,347 \\ \bottomrule
\end{tabular}
\caption{Number of sentence pairs for cLang-8 dataset with candidates. All pairs of data that satisfy the length constraint of 128 are listed.}
\label{table:size_of_training_candidates}
\end{table}

\begin{table*}[t]
\centering
\resizebox{\linewidth}{!}{
\begin{tabular}{|l|lllllllllll|}
\hline
\diagbox{$a_{train}$}{$F_{0.5}$}{\textbf{Threshold($\lambda$)}} & 0 & 0.1 & 0.2 & 0.3 & 0.4 & 0.5 & 0.6 & 0.7 & 0.8 & 0.9 & 1 \\ \hline
0 & \textbf{45.34} & \multicolumn{10}{c|}{\textcolor{gray}{49.36}} \\
5 & 49.14 & 49.10 & 49.64 & 49.86 & \textbf{49.92} & 49.87 & 49.61 & 49.19 & 49.37 & \multicolumn{2}{c|}{\textcolor{gray}{49.36}} \\
10 & 48.84 & 49.50 & 49.62 & 50.09 & \textbf{50.10} & 50.07 & 49.96 & 49.91 & 49.91 & 49.57 & \textcolor{gray}{49.36} \\
20 & 49.13 & 49.42 & 49.74 & 50.08 & \textbf{50.22} & 49.89 & 50.00 & 49.92 & 49.62 & 49.46 & \textcolor{gray}{49.36} \\ \hline
\end{tabular}
}
\caption{Results of tuning $a_{train}$ for BTR. $a_{pred}$ was fixed to 5. The highest $F_{0.5 }$ score on the CoNLL-13 corpus for each $a_{train}$ among different threshold is shown in bold. The scores that were the same as those of the base model ($\lambda=1$) were ignored and \textcolor{gray}{greyed out}.}
\label{table:train_topa_result}
\end{table*}

\begin{table}[t]
\centering
\resizebox{\columnwidth}{!}{
\begin{tabular}{|l|lll|}
\hline
\diagbox{$a_{train}$}{$F_{0.5}$}{\textbf{Threshold($\lambda$)}} & 0 & 0.1 & 0.2, \ldots, 1 \\ \hline
0  & 46.48 & \textbf{49.35} & \textcolor{gray}{49.36} \\
5  & 44.89 & \textbf{49.38} & \textcolor{gray}{49.36} \\
10 & 45.68 & \textbf{49.38} & \textcolor{gray}{49.36} \\
20 & 41.91 & \textbf{49.38} & \textcolor{gray}{49.36} \\ \hline
\end{tabular}
}
\caption{\textcolor{black}{Results of tuning $a_{train}$ for RoBERTa. $a_{pred}$ was fixed to 5. The highest $F_{0.5}$ score on the CoNLL-13 corpus for each $a_{train}$ among different threshold is shown in bold. The scores that were the same as those of the base model ($\lambda=1$) were ignored and \textcolor{gray}{greyed out}.}}
\label{table:train_topa_result_RoBERTa}
\end{table}

\pgfplotsset{width=7cm,compat=1.15}
\pgfplotstableread[row sep=\\,col sep=&]{
    Position & 0candidates & 5candidates & 10candidates & 20candidates & 0candidatesLOSS & 5candidatesLOSS & 10candidatesLOSS & 20candidatesLOSS\\
    1 & 45.02 & 45.22 & 45.37 & 44.13 & 0.080 & 0.028 & 0.021 & 0.019\\
    2 & 46.34 & 44.86 & 44.74 & 42.48 & 0.055 & 0.004 & 0.002 & 0.002\\
    3 & 46.72 & 44.21 & 45.00 & 43.28 & 0.050 & 0.003 & 0.002 & 0.001\\
    4 & 46.64 & 44.67 & 43.93 & 42.63 & 0.048 & 0.003 & 0.002 & 0.001\\
    5 & 46.62 & 44.97 & 44.51 & 42.73 & 0.047 & 0.003 & 0.002 & 0.001\\
    6 & 46.95 & 44.70 & 44.27 & 42.65 & 0.046 & 0.002 & 0.001 & 0.001\\
    7 & 46.87 & 45.12 & 44.41 & 42.41 & 0.045 & 0.002 & 0.001 & 0.001\\
    8 & 47.05 & 44.29 & 44.52 & 43.46 & 0.045 & 0.002 & 0.001 & 0.001\\
    9 & 46.81 & 44.01 & 44.52 & 43.01 & 0.044 & 0.002 & 0.001 & 0.001\\
    10 & 46.97 & 44.07 & 44.09 & 43.03 &0.044 & 0.002 & 0.001 & 0.001\\
    11 & 47.01 & 43.88 & 44.37 & 43.06 &0.044 & 0.002 & 0.001 & 0.001\\
    12 & 47.13 & 43.67 & 44.54 & 43.27 &0.043 & 0.002 & 0.001 & 0.001\\
    13 & 47.01 & 42.79 & 44.81 & 42.83 &0.043 & 0.002 & 0.001 & 0.001\\
    14 & 47.02 & 43.71 & 44.60 & 42.66 &0.043 & 0.002 & 0.001 & 0.001\\
    15 & 46.95 & 43.51 & 44.81 & 42.71 &0.042 & 0.002 & 0.001 & 0.001\\
    }\mydataRobertaFfinetune

\pgfplotsset{width=7cm,compat=1.15}
\pgfplotstableread[row sep=\\,col sep=&]{
    Position & 0candidates & 5candidates & 10candidates & 20candidates & 0candidatesLOSS & 5candidatesLOSS & 10candidatesLOSS & 20candidatesLOSS\\
	1 & 43.89 & 45.43 & 45.59 & 46.58 & 1.105 & 0.694 & 0.456 & 0.323 \\
	2 & 44.40 & 47.54 & 47.45 & 48.02 & 0.190 & 0.527 & 0.338 & 0.239 \\
	3 & 44.78 & 47.86 & 48.64 & 48.51 & 0.151 & 0.473 & 0.304 & 0.216 \\
	4 & 44.79 & 47.92 & 48.56 & 48.67 & 0.129 & 0.441 & 0.283 & 0.202 \\
	5 & 44.92 & 48.55 & 48.38 & 48.91 & 0.115 & 0.418 & 0.268 & 0.191 \\
	6 & 44.78 & 48.47 & 48.57 & 48.87 & 0.106 & 0.399 & 0.256 & 0.183 \\
	7 & 44.89 & 48.44 & 48.74 & 49.05 & 0.099 & 0.383 & 0.245 & 0.175 \\
	8 & 44.78 & 48.58 & 48.51 & 48.65 & 0.093 & 0.369 & 0.236 & 0.169 \\
	9 & 45.14 & 48.69 & 48.62 & 48.69 & 0.088 & 0.356 & 0.227 & 0.162 \\
	10 &45.01 & 48.55 & 48.63 & 48.68 & 0.084 & 0.345 & 0.220 & 0.157 \\
	11 &44.71 & 48.49 & 48.30 & 48.62 & 0.080 & 0.334 & 0.212 & 0.151 \\
	12 &44.83 & 48.54 & 48.53 & 48.44 & 0.077 & 0.324 & 0.205 & 0.147 \\
	13 &44.70 & 48.40 & 48.22 & 48.69 & 0.075 & 0.314 & 0.199 & 0.142 \\
	14 &44.82 & 48.23 & 48.51 & 48.58 & 0.072 & 0.305 & 0.193 & 0.138 \\
	15 &44.89 & 48.23 & 48.40 & 48.20 & 0.069 & 0.297 & 0.187 & 0.134 \\
    }\mydataBTRFfinetune

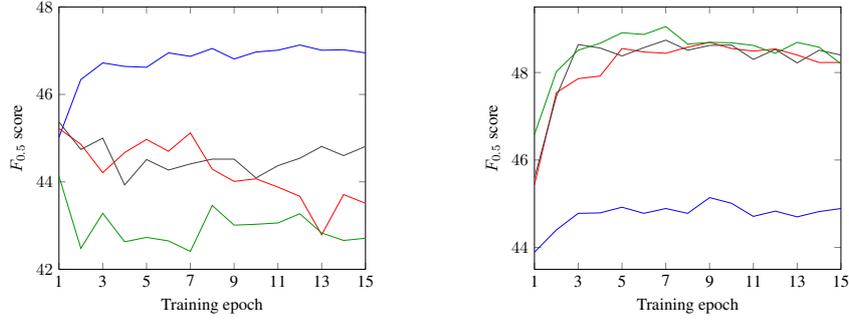
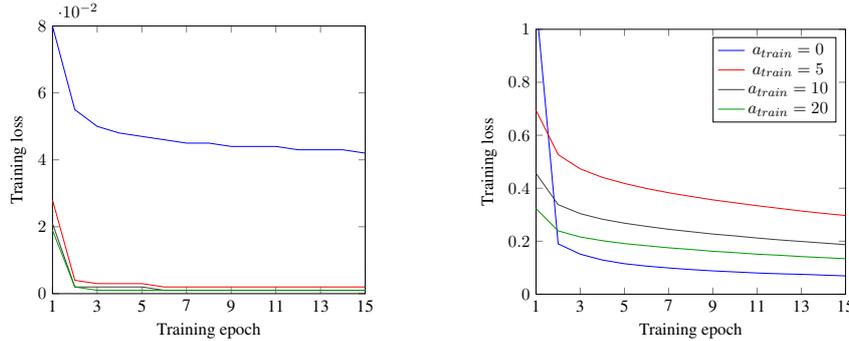
\begin{figure*}[t]
    \centering
    \begin{subfigure}[t]{0.8\columnwidth}  
        \centering
        \begin{adjustbox}{width=0.8\columnwidth}
        \begin{tikzpicture}
        \begin{axis}[
                scale only axis,
                x post scale=1,
                y post scale=1,
                ylabel=$F_{0.5}$ score,
                xlabel=Training epoch,
                ymin=42, ymax=48,
                xmin=1, xmax=15,
                symbolic x coords={1,2,3,4,5,6,7,8,9,10,11,12,13,14,15},
            ]
            \addplot+[sharp plot, mark=none, blue] table[x=Position,y=0candidates]{\mydataRobertaFfinetune};
            \addplot+[sharp plot, mark=none, red] table[x=Position,y=5candidates]{\mydataRobertaFfinetune};
            \addplot+[sharp plot, mark=none, darkgray] table[x=Position,y=10candidates]{\mydataRobertaFfinetune};
            \addplot+[sharp plot, mark=none, green!60!black] table[x=Position,y=20candidates]{\mydataRobertaFfinetune};
        \end{axis}
        \end{tikzpicture}
        \end{adjustbox}
    \caption{$F_{0.5}$ score of RoBERTa on the CoNLL-13 corpus}
    \end{subfigure}
    \begin{subfigure}[t]{0.8\columnwidth}
        \centering
        \begin{adjustbox}{width=0.8\columnwidth}
        \begin{tikzpicture}
            \begin{axis}[
                scale only axis,
                x post scale=1,
                y post scale=1,
                ylabel=$F_{0.5}$ score,
                xlabel=Training epoch,
                ymin=43.5, ymax=49.5,
                xmin=1, xmax=15,
                symbolic x coords={1,2,3,4,5,6,7,8,9,10,11,12,13,14,15},
            ]
            \addplot+[sharp plot, mark=none, blue] table[x=Position,y=0candidates]{\mydataBTRFfinetune};
            \addplot+[sharp plot, mark=none, red] table[x=Position,y=5candidates]{\mydataBTRFfinetune};
            \addplot+[sharp plot, mark=none, darkgray] table[x=Position,y=10candidates]{\mydataBTRFfinetune};
            \addplot+[sharp plot, mark=none, green!60!black] table[x=Position,y=20candidates]{\mydataBTRFfinetune};
        \end{axis}
        \end{tikzpicture}
        \end{adjustbox}
    \caption{$F_{0.5}$ score of BTR on the CoNLL-13 corpus}
    \end{subfigure}
    
    \begin{subfigure}[t]{0.8\columnwidth}  
        \centering
        \begin{adjustbox}{width=0.8\columnwidth}
        \begin{tikzpicture}
        \begin{axis}[
                scale only axis,
                x post scale=1,
                y post scale=1,
                ylabel=Training loss,
                xlabel=Training epoch,
                ymin=0, ymax=0.08,
                xmin=1, xmax=15,
                symbolic x coords={1,2,3,4,5,6,7,8,9,10,11,12,13,14,15},
            ]
            \addplot+[sharp plot, mark=none, blue] table[x=Position,y=0candidatesLOSS]{\mydataRobertaFfinetune};
            \addplot+[sharp plot, mark=none, red] table[x=Position,y=5candidatesLOSS]{\mydataRobertaFfinetune};
            \addplot+[sharp plot, mark=none, darkgray] table[x=Position,y=10candidatesLOSS]{\mydataRobertaFfinetune};
            \addplot+[sharp plot, mark=none, green!60!black] table[x=Position,y=20candidatesLOSS]{\mydataRobertaFfinetune};
        \end{axis}
        \end{tikzpicture}
        \end{adjustbox}
    \caption{Training loss of RoBERTa}
    \end{subfigure}
    \begin{subfigure}[t]{0.8\columnwidth}
        \centering
        \begin{adjustbox}{width=0.82\columnwidth}
        \begin{tikzpicture}
            \begin{axis}[
                scale only axis,
                legend pos= north east,
                x post scale=1,
                y post scale=1,
                ylabel=Training loss,
                xlabel=Training epoch,
                ymin=0, ymax=1,
                xmin=1, xmax=15,
                symbolic x coords={1,2,3,4,5,6,7,8,9,10,11,12,13,14,15},
            ]
            \addlegendimage{/pgfplots/refstyle=plot_BTR_0_cand_loss}\addlegendentry{$a_{train}=0$}
            \addlegendimage{/pgfplots/refstyle=plot_BTR_5_cand_loss}\addlegendentry{$a_{train}=5$}
            \addlegendimage{/pgfplots/refstyle=plot_BTR_10_cand_loss}\addlegendentry{$a_{train}=10$}
            \addlegendimage{/pgfplots/refstyle=plot_BTR_20_cand_loss}\addlegendentry{$a_{train}=20$}
            \addplot+[sharp plot, mark=none, blue] table[x=Position,y=0candidatesLOSS]{\mydataBTRFfinetune};\label{plot_BTR_0_cand_loss}
            \addplot+[sharp plot, mark=none, red] table[x=Position,y=5candidatesLOSS]{\mydataBTRFfinetune};\label{plot_BTR_5_cand_loss}
            \addplot+[sharp plot, mark=none, darkgray] table[x=Position,y=10candidatesLOSS]{\mydataBTRFfinetune};\label{plot_BTR_10_cand_loss}
            \addplot+[sharp plot, mark=none, green!60!black] table[x=Position,y=20candidatesLOSS]{\mydataBTRFfinetune};\label{plot_BTR_20_cand_loss}
        \end{axis}
        \end{tikzpicture}
        \end{adjustbox}
    \caption{Training loss of BTR}
    \end{subfigure}
\caption{Performances of BTR and RoBERTa with various $a_{train}$ without $\lambda$ during fine-tuning. $a_{pred}$ was fixed to 5 with candidates from T5GEC. Both $F_{0.5}$ score and training loss were averaged over the four trials.}
\label{fig:diff_atrain}
\end{figure*}

\begin{table*}[ht]
\centering
\resizebox{\linewidth}{!}{
\begin{tabular}{|l|lllllllllll|}
\hline
\diagbox{$a_{pred}$}{$F_{0.5}$}{\textbf{Threshold($\lambda$)}} & 0 & 0.1 & 0.2 & 0.3 & 0.4 & 0.5 & 0.6 & 0.7 & 0.8 & 0.9 & 1 \\ \hline
5 & 49.13 & 49.42 & 49.74 & 50.08 & \textbf{50.22} & 49.89 & 50.00 & 49.92 & 49.62 & 49.46 & \textcolor{gray}{49.36} \\
10 & 48.92 & 49.34 & \textbf{50.01} & 49.85 & 49.85 & 49.51 & 49.71 & 49.62 & 49.49 & 49.39 & \textcolor{gray}{49.40} \\
15 & 48.91 & 49.22 & \textbf{49.65} & 49.36 & 49.21 & 49.18 & 49.04 & 49.08 & 48.90 & 48.92 & \textcolor{gray}{48.88} \\
20 & 36.50 & 36.83 & 38.21 & 38.85 & 40.24 & 41.84 & 43.11 & 44.41 & 45.65 & \textbf{46.87} & \textcolor{gray}{49.40} \\ \hline
\end{tabular}
}
\caption{Results of tuning $a_{pred}$ for BTR. The highest $F_{0.5}$ score on the CoNLL-13 corpus for each $a_{pred}$ among different threshold is shown in bold. The scores that were same as those of the base model ($\lambda=1$) were ignored and \textcolor{gray}{greyed out}.}
\label{table:eval_topa_result_BTR}
\end{table*}

\begin{table}[t]
\centering
\resizebox{1\columnwidth}{!}{
\begin{tabular}{|l|ccc|}
\hline
\diagbox{$a_{pred}$}{$F_{0.5}$}{\textbf{Threshold($\lambda$)}} & 0 & 0.1 & 0.2, \ldots, 1 \\ \hline
5 & 46.48 & \textbf{49.35} & \textcolor{gray}{49.36} \\
10 & \textbf{46.08} & \multicolumn{2}{c|}{\textcolor{gray}{49.40}}  \\
15 & \textbf{45.04} & \multicolumn{2}{c|}{\textcolor{gray}{48.88}} \\
20 & \textbf{44.28} & \multicolumn{2}{c|}{\textcolor{gray}{49.40}}  \\ \hline
\end{tabular}
}
\caption{Results of tuning $a_{pred}$ for RoBERTa. The highest $F_{0.5}$ score on the CoNLL-13 corpus for each $a_{pred}$ among different threshold is shown in bold. The scores that were same as those of the base model ($\lambda=1$) were ignored and \textcolor{gray}{greyed out}.}
\label{table:eval_topa_result_RoBERTa}
\end{table}

\begin{table}[ht]
\centering
\resizebox{0.7\columnwidth}{!}{
\begin{tabular}{lrcc}
\toprule
\textbf{Dataset} & $a_{pred}$ & \textbf{R2L} & \textbf{BERT} \\ \midrule
& 5  & \textbf{50.02} & \textbf{42.44} \\ 
& 10  & 49.94 & 40.53 \\ 
& 15  & 49.85 & 39.98 \\
\multirow{-3}{*}{CoNLL-13} & 20  & 39.81 & 39.37\\ \bottomrule
\end{tabular}
}
\caption{Results of tuning $a_{pred}$ for R2L and BERT. The highest $F_{0.5}$ score on the CoNLL-13 corpus for each reranker among different $a_{pred}$ is shown in bold.}
\label{table:eval_topa_result_R2L}
\end{table}

\begin{table*}[ht]
\centering
\resizebox{0.9\linewidth}{!}{
\begin{tabular}{lccccccccccc}
\toprule
\multirow{2}{*}{\textbf{Model}} & \multicolumn{11}{c}{$\lambda$} \\ \cmidrule(lr){2-12}
 & 0 & 0.1 & 0.2 & 0.3 & 0.4 & 0.5 & 0.6 & 0.7 & 0.8 & 0.9 & 1 \\ \midrule
RoBERTa & 47.90 & \textbf{50.76} &  \multicolumn{9}{c}{\textcolor{gray}{50.79}} \\
BTR & 49.44 & 50.17 & 50.00 & 49.98 & 49.98 & 50.58 & 50.47 & 50.92 & \textbf{51.00} & 50.82 & \textcolor{gray}{50.79} \\ \bottomrule
\end{tabular}
}
\caption{Results of RoBERTa and BTR on the CoNLL-13 corpus with candidates generated by T5GEC (large). The scores that were the same as those of the base model ($\lambda=1$) were ignored and \textcolor{gray}{greyed out}.}
\label{table:valid_results_large}
\end{table*}

Tables \ref{table:train_topa_result} and \ref{table:eval_topa_result_BTR} also show the performances of the BTR concerning $\lambda$ on the CoNLL-13 corpus with candidates generated by T5GEC.
Without using any candidate for training, the BTR($\lambda=0$) could achieve the highest $F_{0.5}$ score. When using 20 candidates for training, the BTR ($\lambda=0.4$) achieved the highest $F_{0.5}$ score of 50.22.
Table \ref{table:valid_results_large} shows the BTR ($a_{train}=20, \lambda=0.8$) achieved the highest $F_{0.5}$ score on the CoNLL-13 dataset with the candidates generated by T5GEC(large).
Thus, our tuned $\lambda$ for the BTR was set to 0.2 when $a_{train}=0$. When $a_{train}=20$, $\lambda$ was set to 0.4 and 0.8 for the candidates generated by T5GEC and T5GEC(large), respectively. Similarly, when $a_{train}=0$, our tuned $\lambda$ for RoBERTa was set to 0.1 for the two versions of candidates.

\section{Mean and Standard Deviation}
\label{section:appendix_mean_std}
\begin{table*}[ht]
\centering
\resizebox{0.8\linewidth}{!}{
\begin{tabular}{lcccc}
\toprule
\textbf{Model} & \textbf{CoNLL-13} & \textbf{CoNLL-14} & \textbf{BEA} & \textbf{JFLEG}\\ \midrule
R2L & 50.02 $\pm$ 0.15 & 64.96 $\pm$  0.10 & \textbf{71.42} $\pm$  0.05 & 59.09 $\pm$  0.19 \\
RoBERTa w/o $a_{train}$ & 48.66 $\pm$ 1.20 & 63.96 $\pm$ 1.90 & 68.90 $\pm$ 2.88 & 58.68 $\pm$ 0.66\\
BTR & \textbf{50.05} $\pm$  0.07 & \textbf{65.29} $\pm$ 0.19 & 70.84 $\pm$  0.05 & \textbf{59.12} $\pm$  0.07 \\ \bottomrule
\end{tabular}
}
\caption{The mean $\pm$ std results on each dataset with candidates from T5GEC. Bold scores are the highest mean for each dataset.}
\label{table:mean_std}
\end{table*}

We list the mean and standard deviation of R2L, RoBERTa, and the BTR over the four trails on each dataset in Table \ref{table:mean_std}.

\section{Detailed Results on BEA Test}
\label{sec:appendix_cefr_results}

\begin{table}[t]
\centering
\begin{tabular}{lcr}
\toprule
\textbf{Dataset} & \textbf{Level} & \textbf{\# of data (pairs)} \\ \midrule
  & A & 1,107 \\
  & B & 1,330 \\
  & C & 1,010 \\
\multirow{-4}{*}{BEA}  & N & 1,030 \\ \bottomrule
\end{tabular}
\caption{Dataset size of the BEA test. Each sentence in the BEA test set is classified into either A (beginner), B (intermediate), C (advanced), or N (native) corresponding to the CEFR level.}
\label{table:beatest_data_distribution}
\end{table}

\begin{table}[t]
\resizebox{\columnwidth}{!}{
\begin{tabular}{lcllll}
\toprule
 \textbf{Model} & \textbf{Level} & \rotatebox{60}{\textbf{Missing}} & \rotatebox{60}{\textbf{Replacement}} & \rotatebox{60}{\textbf{Unnecessary}} & \textbf{All} \\ \midrule
 & A & 62.30 & \textbf{69.92} & 73.74 & 68.40 \\
 & B & 73.99 & 67.94 & 78.26 & 70.93 \\
 & C & 78.54 & 71.51 & 85.16 & 75.54 \\
 & N & 80.66 & \textbf{69.48} & 53.78 & 71.36 \\
\multirow{-5}{*}{T5GEC} & All & 71.23 & 69.47 & 74.30 & 70.51 \\ \midrule
 & A & \textbf{63.65} & 69.86 & \textbf{74.40} & \textbf{68.76} \\
 & B & \textbf{74.81} & \textbf{68.94} & \textbf{79.01} & \textbf{71.84} \\
 & C & \textbf{81.85} & \textbf{72.61} & \textbf{86.00} & \textbf{77.36} \\
 & N & \textbf{83.17} & 68.23 & \textbf{57.88} & \textbf{72.01} \\
\multirow{-5}{*}{BTR ($\lambda=0.4$)} & all & \textbf{72.91} & \textbf{69.69} & \textbf{75.69} & \textbf{71.27} \\ \bottomrule
\end{tabular}
}
\caption{Results for each operation type with classified CEFR levels on the BEA test set with candidates from T5GEC. Edit operations are divided into \emph{Missing}, \emph{Replacement}, and \emph{Unnecessary} corresponding to inserting, substituting, and deleting tokens, respectively. Bold scores are the highest for each operation with the corresponding level.}
\label{table:Level_Error_results}
\end{table}

\begin{table}[t]
\centering
\resizebox{1\columnwidth}{!}{
\begin{tabular}{llllll}
\toprule
\textbf{Model} & \textbf{PUNCT} & \textbf{DET} & \textbf{PREP} & \textbf{ORTH} & \textbf{SPELL} \\ \midrule
T5GEC & 74.62 & 77.57 & 73.33 & 70.32 & 78.38 \\
BTR ($\lambda=0.4$) & \textbf{75.73} & \textbf{79.08} & \textbf{73.77} & \textbf{70.72} & \textbf{78.87} \\ \bottomrule
\end{tabular}
}
\caption{Results for the top five error types on the BEA test set. Bold scores are the highest for each error type.}
\label{table:error_type_result}
\end{table}

The distribution of the BEA test set with respect to the CEFR level is shown in Table \ref{table:beatest_data_distribution}.

The BTR ($\lambda=0.4$) achieved an $F_{0.5}$ score of 71.27 on the BEA test set, as shown in Table \ref{table:Level_Error_results}. Compared with A (beginner) level sentences, the BTR was more effective for B (intermediate), C (advanced), and N (native) level sentences. As shown in Table \ref{table:error_type_result}, the BTR ($\lambda=0.4$) improved T5GEC for all top-5 error types.
Furthermore, the BTR ($\lambda=0.4$) could effectively handle \emph{Missing} and \emph{Unnecessary} tokens but not \emph{Replacement} for the native sentences. 
It was more difficult to correct the \emph{Replacement} and \emph{Unnecessary} operations in the native sentences for both models compared with the advanced sentences. 
This may be because the writing style of native speakers is more natural and difficult to correct with limited training data, whereas language learners may tend to use a formal language to make the correction easier.

\section{Relation Between $a$, $\lambda$, and BTR Performance}
\label{sec:appendix_relation_between_lambda_pre_rec}

\begin{table*}[t]
\centering
\resizebox{\linewidth}{!}{
\begin{tabular}{|l|ccccccccccc|}
\hline
\diagbox{$a_{train}$, $a_{pred}$}{$F_{0.5}$}{\textbf{Threshold($\lambda$)}} & 0 & 0.1 & 0.2 & 0.3 & 0.4 & 0.5 & 0.6 & 0.7 & 0.8 & 0.9 & 1 \\ \hline
0, 5  & 47.44 & \textbf{52.83} & 52.77 & 52.52 & \multicolumn{7}{c|}{\textcolor{gray}{52.51}} \\
0, 10 & 44.93 & \textbf{52.65} & \multicolumn{9}{c|}{\textcolor{gray}{52.37}} \\
0, 15 & 43.74 & \textbf{52.46} & \textbf{52.46} & \textbf{52.46} & \textbf{52.46} & \textbf{52.46} & \textbf{52.46} & \textbf{52.46} & \textbf{52.46} & \multicolumn{2}{c|}{\textcolor{gray}{52.45}} \\
0, 20 & 31.01 & 51.86 & 52.21 & 52.33 & 52.38 & 52.41 & 52.42 & 52.44 & \textbf{52.45} & \textbf{52.45} & \textcolor{gray}{52.47} \\ \hline
5, 5  & 52.51 & 53.22 & \textbf{53.49} & 53.41 & 53.42 & 53.45 & 53.40 & 53.28 & 53.17 & 53.06 & \textcolor{gray}{52.51} \\
5, 10 & 51.21 & 53.19 & 53.52 & 53.41 & 53.40 & \textbf{53.56} & 53.34 & 53.21 & 53.15 & 53.04 & \textcolor{gray}{52.37} \\
5, 15 & 50.68 & 53.11 & 53.37 & 53.45 & \textbf{53.54} & 53.46 & 53.30 & 53.23 & 53.20 & 53.10 & \textcolor{gray}{52.45} \\
5, 20 & 30.44 & 32.42 & 34.86 & 36.39 & 38.28 & 41.33 & 42.73 & 43.90 & 45.16 & \textbf{46.67} & \textcolor{gray}{52.47} \\ \hline
10, 5 & 53.47 & 53.95 & \textbf{54.04} & 53.95 & 53.85 & 53.68 & 53.64 & 53.38 & 53.09 & 53.01 & \textcolor{gray}{52.51} \\
10,10 & 52.51 & 53.21 & \textbf{53.99} & 53.87 & 53.70 & 53.73 & 54.49 & 53.30 & 53.02 & 52.99 & \textcolor{gray}{52.37} \\
10,15 & 52.05 & 53.97 & \textbf{54.01} & 53.64 & 53.66 & 53.63 & 53.44 & 53.26 & 53.03 & 53.05 & \textcolor{gray}{52.45} \\
10,20 & 30.03 & 31.55 & 32.76 & 34.12 & 36.07 & 39.25 & 40.89 & 42.50 & 43.68 & \textbf{45.45} & \textcolor{gray}{52.47} \\ \hline
20, 5 & 53.26 & \textbf{53.87} & 53.85 & 53.75 & 53.79 & 53.77 & 53.70 & 53.50 & 53.31 & 52.98 & \textcolor{gray}{52.51} \\
20,10 & 52.37 & 53.15 & 53.75 & 53.77 & \textbf{53.91} & 53.83 & 53.54 & 53.44 & 53.24 & 53.15 & \textcolor{gray}{52.37} \\
20,15 & 52.29 & 53.69 & 53.82 & \textbf{53.85} & 53.84 & 53.64 & 53.46 & 53.33 & 53.21 & 53.21 & \textcolor{gray}{52.45} \\
20,20 & 29.68 & 31.03 & 32.11 & 33.47 & 35.15 & 38.52 & 39.99 & 41.64 & 43.25 & \textbf{44.95} & \textcolor{gray}{52.47} \\ \hline
\end{tabular}
}
\caption{\textcolor{black}{Results of tuning $a_{train}$ and $a_{pred}$ for BTR on the BEA dev set. The highest $F_{0.5}$ score for each pair of $a_{train}$ and $a_{pred}$ among different threshold is shown in bold. The scores that were the same as those of the base model ($\lambda=1$) were ignored and \textcolor{gray}{greyed out}.}}
\label{table:tuning_a_for_BTR_on_BEA}
\end{table*}

\begin{table*}[t]
\centering
\resizebox{\linewidth}{!}{
\begin{tabular}{|l|ccccccccccc|}
\hline
\diagbox{$a_{train}$, $a_{pred}$}{\textbf{GLEU}}{\textbf{Threshold($\lambda$)}} & 0 & 0.1 & 0.2 & 0.3 & 0.4 & 0.5 & 0.6 & 0.7 & 0.8 & 0.9 & 1 \\ \hline
0, 5  & \textbf{52.43} & \multicolumn{10}{c|}{\textcolor{gray}{53.25}} \\
0, 10 & \textbf{51.91} & \multicolumn{10}{c|}{\textcolor{gray}{53.25}} \\
0, 15 & \textbf{51.36} & \multicolumn{10}{c|}{\textcolor{gray}{53.25}} \\
0, 20 & 44.59 & \textbf{52.97} & \multicolumn{9}{c|}{\textcolor{gray}{53.25}} \\ \hline
5, 5  & 54.35 & \textbf{54.37} & 54.12 & 54.05 & 53.81 & 53.67 & 53.42 & 53.32 & 53.20 & 53.22 & \textcolor{gray}{53.25} \\
5, 10 & \textbf{54.41} & 54.34 & 54.05 & 53.68 & 53.48 & 53.31 & 53.30 & 53.26 & 53.26 & \multicolumn{2}{c|}{\textcolor{gray}{53.25}} \\
5, 15 & \textbf{54.46} & 54.44 & 53.88 & 53.43 & 53.33 & 53.29 & 53.22 & 53.22 & 53.26 & 53.20 & \textcolor{gray}{53.25} \\
5, 20 & 44.42 & 44.99 & 45.82 & 46.32 & 46.80 & 47.43 & 47.73 & 48.13 & 48.98 & \textbf{49.37} & \textcolor{gray}{53.25} \\ \hline
10, 5 & 54.15 & \textbf{54.23} & 53.99 & 53.88 & 53.69 & 53.51 & 53.37 & 53.28 & 53.24 & 53.23 & \textcolor{gray}{53.25} \\
10,10 & \textbf{54.23} & 54.20 & 53.93 & 53.73 & 53.54 & 53.37 & 53.29 & 53.24 & 53.24 & 53.23 & \textcolor{gray}{53.25} \\
10,15 & \textbf{54.29} & 54.16 & 53.89 & 53.57 & 53.48 & 53.33 & 53.26 & 53.19 & 53.22 & 53.23 & \textcolor{gray}{53.25} \\
10,20 & 44.22 & 44.71 & 45.29 & 45.70 & 46.20 & 47.21 & 47.45 & 47.98 & 48.52 & \textbf{49.07} & \textcolor{gray}{53.25} \\ \hline
20, 5 & \textbf{53.92} & 53.87 & 53.85 & 53.68 & 53.60 & 53.49 & 53.50 & 53.38 & 53.25 & 53.26 & \textcolor{gray}{53.25} \\
20,10 & \textbf{53.92} & 53.88 & 53.65 & 53.54 & 53.53 & 53.42 & 53.32 & 53.22 & 53.23 & 53.26 & \textcolor{gray}{53.25} \\
20,15 & \textbf{54.12} & 53.89 & 53.61 & 53.42 & 53.42 & 53.38 & 53.28 & 53.23 & 53.19 & 53.22 & \textcolor{gray}{53.25} \\
20,20 & 44.37 & 44.79 & 45.22 & 45.56 & 45.98 & 46.93 & 47.36 & 47.83 & 48.48 & \textbf{49.08} & \textcolor{gray}{53.25} \\ \hline
\end{tabular}
}
\caption{\textcolor{black}{Results of tuning $a_{train}$ and $a_{pred}$ for BTR on the JFLEG dev set. The highest GLEU score for each pair of $a_{train}$ and $a_{pred}$ among different threshold is shown in bold. The scores that were the same as those of the base model ($\lambda=1$) were ignored and \textcolor{gray}{greyed out}.}}
\label{table:tuning_a_for_BTR_on_JFLEG}
\end{table*}

\begin{table}[ht]
\centering
\resizebox{1\columnwidth}{!}{
\begin{tabular}{llllll}
\toprule
\textbf{Tuned on corpus} & \textbf{$a_{train}$} & \textbf{$a_{pred}$} & \textbf{$\lambda$} & \textbf{BEA} & \textbf{JFLEG}   \\ \midrule
CoNLL-13 & 20          & 5          & 0.4       & \textbf{71.27} & 59.17      \\
BEA dev &  10         &  5         &  0.2 & 71.12	 & -     \\ 
JFLEG dev & 5           & 15         & 0         & -     & \textbf{60.14} \\ \bottomrule
\end{tabular}
}
\caption{\textcolor{black}{Results for BTR on the BEA and JFLEG test sets with tuned hyperparameters.}}
\label{table:tuned_a_lambda_for_BTR_onBEAJFLEG}
\end{table}

The BEA and JFLEG corpus also provide a dev set with 4384 and 754 sentences for validation, respectively. To determine the optimal $a_{train}$, $a_{pred}$, and $\lambda$ for the BTR listed in Table \ref{table:train_topa_result} on these two datasets, we re-evaluated the performances of the BTR on the corresponding dev sets. Tables \ref{table:tuning_a_for_BTR_on_BEA} and \ref{table:tuning_a_for_BTR_on_JFLEG} show the results on the BEA and JFLEG dev sets, respectively. 
\textcolor{black}{On the BEA dev set, the highest $F_{0.5}$ score of 54.04 was achieved with $a_{train} = 10$, $a_{pred}=5$, and $\lambda=0.2$.}
\textcolor{black}{On the JFLEG dev set, the highest GLEU score of 54.46 was achieved with $a_{train} = 5$, $a_{pred}=15$, and $\lambda=0$.}
These results demonstrate the differences in evaluating the minimal edit and fluency for grammar corrections. 
Given the previous $a_{train}, a_{pred}$ and $\lambda$, we re-evaluated the BTR on the BEA and JFLEG test sets. Table \ref{table:tuned_a_lambda_for_BTR_onBEAJFLEG} lists the results. 
\textcolor{black}{Tuning hyperparameters on the JFLEG dev set led to a higher GLEU score of 60.14 on the JFLEG test set, compared to the tuned hyperparameters on the CoNLL-13 set. However, tuning hyperparameters on the BEA dev set resulted in a lower $F_{0.5}$ score of 71.12 on the BEA test set, compared to the tuned hyperparameters on the CoNLL-13 set.}

To investigate the effectiveness of $\lambda$, \textcolor{black}{i.e., the parameter that balances the trade-off between acceptance rate and quality of grammatical corrections}, we analyzed the relationship between $\lambda$ and the corresponding precision, recall, and GLEU scores. Figures \ref{fig:all_versus_lambda_conll13} and \ref{fig:all_versus_lambda_conll14} show the performance of the BTR ($a_{train}=20, a_{pred}=5$) on the CoNLL-13 and 14 corpus, respectively. With $\lambda$ increasing, the acceptance rate, \textcolor{black}{i.e., the percentage of suggestions that the BTR accepts,} decreases while the precision and recall for the \emph{Accept} suggestions increases. This demonstrates our assumption in Section \ref{sec:inference} that \textcolor{black}{the value of} $f(\bm{y}|\bm{x})$ indicates the confidence of the BTR, and the confident suggestions tended to be more grammatical, while the unconfident ones tended to be less grammatical than the original selections. 
As for the whole corpus, when $\lambda=0.7$, this BTR achieved lower precision and recall score than $\lambda=0.4$ due to the limited amount of $Accept$ suggestions. Figures \ref{fig:all_versus_lambda_beadev} and \ref{fig:all_versus_lambda_beatest} show the performance of BTR ($a_{train}=10, a_{pred}=5$) on the BEA dev and test corpus, respectively. In Figure \ref{fig:all_versus_lambda_beadev}, the BTR shows a similar performance to that on the CoNLL-13 and 14 that, 
\textcolor{black}{where a larger $\lambda$ leads to higher precision and recall for $Accept$ suggestions.}
However, the performance over the whole corpus also depends on the acceptance rate. Differently, as shown in Figures \ref{fig:all_versus_lambda_jflegdev} and \ref{fig:all_versus_lambda_jflegtest}, the experimental results of the BTR ($a_{train}=5, a_{pred}=15$) on the JFLEG corpus achieved the highest GLEU score for the whole corpus when $\lambda \leq 0.1$. This may be because using $a_{pred}=15$ makes a flatter probability than $a_{pred}=5$ as shown in Figures \ref{fig:probability_distribution_of_ranking_BEA} and \ref{fig:probability_distribution_of_ranking_JFLEG}. Besides, recognizing the fluency of a sentence by the BTR may be easier than recognizing the minimal edit of corrections.

\pgfplotsset{width=7cm,compat=1.15}
\pgfplotstableread[row sep=\\,col sep=&]{
    Threshold & Acceptrate & Precision & Recall & Precisionforaccept & Recallforaccept &Precisionybase & Recallybase\\
    0   &34.76 & 58.10 & 30.37 & 53.62 & 31.88 & 55.36 & 30.37 \\
    0.1 &27.66 & 58.57 & 30.43 & 54.07 & 31.49 & 55.28 & 29.32 \\
    0.2 &22.30 & 59.00 & 30.57 & 54.55 & 31.8  & 54.70 & 28.5 \\
    0.3 &17.31 & 59.53 & 30.63 & 55.17 & 32.56 & 53.23 & 27.95 \\
    0.4 &12.96 & 59.87 & 30.54 & 54.79 & 31.62 & 50.00 & 25.71 \\
    0.5 & 9.12 & 59.64 & 30.17 & 55.10 & 30.42 & 50.28 & 25.35 \\
    0.6 & 6.81 & 59.80 & 30.20 & 59.59 & 32.95 & 51.52 & 25.76 \\
    0.7 & 4.34 & 59.77 & 30.08 & 61.96 & 34.76 & 50.00 & 25.61 \\
    0.8 & 2.17 & 59.45 & 29.85 & 56.82 & 33.33 & 46.15 & 24.00 \\
    0.9 & 1.01 & 59.29 & 29.74 & 47.83 & 29.73 & 40.00 & 21.62 \\
    1   & 0.00 & 59.19 & 29.65 & 0     & 0   & & \\
    }\mydatac

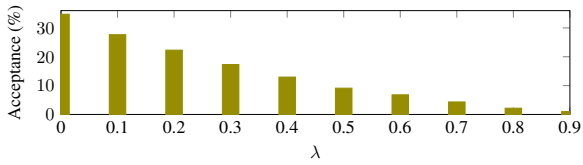
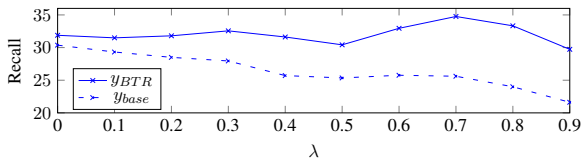
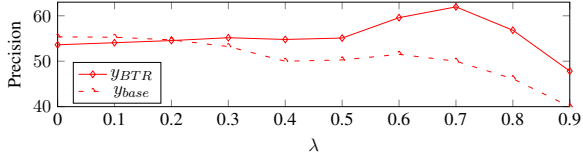
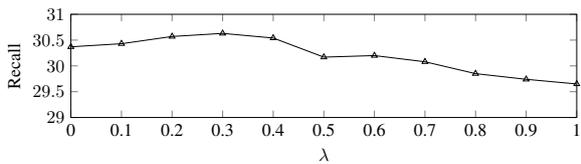
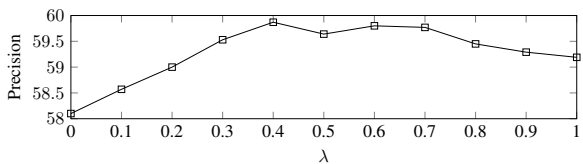
\begin{figure}[ht]
    \begin{subfigure}[t]{1\columnwidth}
        \begin{adjustbox}{width=1\columnwidth}
        \begin{tikzpicture}
            \begin{axis}[
                    x post scale=2,
                    y post scale=0.5,
                    ylabel=Acceptance (\%),
                    xlabel=$\lambda$,
                    ymin=0, ymax=36,
                    xmin=0, xmax=0.9,
                    symbolic x coords={0,0.1,0.2,0.3,0.4,0.5,0.6,0.7,0.8,0.9,1},
                ]
                \addplot [olive, ybar, ybar legend, fill=olive] table[x=Threshold, y=Acceptrate]{\mydatac}; 
            \end{axis}
        \end{tikzpicture}
        \end{adjustbox}
        \caption{Accept rate versus $\lambda$} 
    \end{subfigure}
    
    \begin{subfigure}[t]{1\columnwidth}
        \begin{adjustbox}{width=1\columnwidth}
        \begin{tikzpicture}
            \begin{axis}[
                    legend pos=south west,
                    x post scale=2,
                    y post scale=0.5,
                    ylabel=Recall,
                    xlabel=$\lambda$,
                    ymin=20, ymax=36,
                    xmin=0, xmax=0.9,
                    symbolic x coords={0,0.1,0.2,0.3,0.4,0.5,0.6,0.7,0.8,0.9,1},
                ]
                \addlegendimage{/pgfplots/refstyle=plot_con13_recall_ybtr}\addlegendentry{$y_{BTR}$}
                \addlegendimage{/pgfplots/refstyle=plot_con13_recall_ybase}\addlegendentry{$y_{base}$}
                \addplot+[sharp plot, mark=x, blue] table[x=Threshold,y=Recallforaccept]{\mydatac};\label{plot_con13_recall_ybtr}
                \addplot+[sharp plot, loosely dashed, mark=x, blue] table[x=Threshold,y=Recallybase]{\mydatac};\label{plot_con13_recall_ybase}
            \end{axis}
        \end{tikzpicture}
        \end{adjustbox}
        \caption{Recall for accepted suggestions versus $\lambda$} 
    \end{subfigure}

    \begin{subfigure}[t]{1\columnwidth}
        \begin{adjustbox}{width=1\columnwidth}
        \begin{tikzpicture}
            \begin{axis}[
                    legend pos=south west,
                    x post scale=2,
                    y post scale=0.5,
                    ylabel=Precision,
                    xlabel=$\lambda$,
                    ymin=40, ymax=63,
                    xmin=0, xmax=0.9,
                    symbolic x coords={0,0.1,0.2,0.3,0.4,0.5,0.6,0.7,0.8,0.9,1},
                ]
                \addlegendimage{/pgfplots/refstyle=plot_con13_precision_ybtr}\addlegendentry{$y_{BTR}$}
                \addlegendimage{/pgfplots/refstyle=plot_con13_precision_ybase}\addlegendentry{$y_{base}$}
                \addplot+[sharp plot, mark=diamond, red] table[x=Threshold,y=Precisionforaccept]{\mydatac};\label{plot_con13_precision_ybtr}
                \addplot+[sharp plot, loosely dashed, mark=diamond, red] table[x=Threshold,y=Precisionybase]{\mydatac};\label{plot_con13_precision_ybase}
            \end{axis}
        \end{tikzpicture}
        \end{adjustbox}
        \caption{Precision for accepted suggestions versus $\lambda$} 
    \end{subfigure}

    \begin{subfigure}[t]{1\columnwidth}
        \begin{adjustbox}{width=1\columnwidth}
        \begin{tikzpicture}
            \begin{axis}[
                    x post scale=2,
                    y post scale=0.5,
                    ylabel=Recall,
                    xlabel=$\lambda$,
                    ymin=29, ymax=31,
                    xmin=0, xmax=1,
                    symbolic x coords={0,0.1,0.2,0.3,0.4,0.5,0.6,0.7,0.8,0.9,1},
                ]
                \addplot+[sharp plot, mark=triangle, black] table[x=Threshold,y=Recall]{\mydatac};
            \end{axis}
        \end{tikzpicture}
        \end{adjustbox}
        \caption{Recall over the whole corpus versus $\lambda$} 
    \end{subfigure}

    \begin{subfigure}[t]{1\columnwidth}
        \begin{adjustbox}{width=1\columnwidth}
        \begin{tikzpicture}
            \begin{axis}[
                    x post scale=2,
                    y post scale=0.5,
                    ylabel=Precision,
                    xlabel=$\lambda$,
                    ymin=58, ymax=60,
                    xmin=0, xmax=1,
                    symbolic x coords={0,0.1,0.2,0.3,0.4,0.5,0.6,0.7,0.8,0.9,1},
                ]
                \addplot+[sharp plot, mark=square, black] table[x=Threshold,y=Precision]{\mydatac};      
            \end{axis}
        \end{tikzpicture}
        \end{adjustbox}
        \caption{Precision over the whole corpus versus $\lambda$} 
    \end{subfigure}
    \caption{Precision and recall of BTR ($a_{train}=20, a_{pred}=5$) with respect to different $\lambda$ on the CoNLL-13 set.} 
    \label{fig:all_versus_lambda_conll13}
\end{figure}

\pgfplotsset{width=7cm,compat=1.15}
\pgfplotstableread[row sep=\\,col sep=&]{
    Threshold & Acceptrate & Precision & Recall & Precisionforaccept & Recallforaccept &Precisionybase & Recallybase \\
    0   & 33.61   & 69.52 & 48.07 & 63.62 & 46.54 & 66.81 & 47.15\\
    0.1 & 26.52   & 70.20 & 48.41 & 65.28 & 47.72 & 67.90 & 47.70 \\
    0.2 & 21.34   & 70.91 & 48.40 & 65.87 & 47.10 & 67.04 & 47.14 \\
    0.3 & 16.69   & 71.19 & 48.45 & 66.67 & 46.95 & 67.01 & 46.74 \\
    0.4 & 12.50   & 71.62 & 48.74 & 69.07 & 49.39 & 66.78 & 47.22 \\
    0.5 &  8.84   & 71.62 & 48.80 & 71.14 & 53.16 & 67.88 & 48.89 \\
    0.6 &  6.78   & 71.75 & 48.72 & 72.44 & 53.55 & 66.67 & 49.02 \\
    0.7 &  4.42   & 71.48 & 48.59 & 72.28 & 54.89 & 68.42 & 50.39 \\
    0.8 &  2.52   & 71.16 & 48.31 & 71.70 & 57.58 & 75.47 & 59.70\\
    0.9 &  1.14   & 71.03 & 48.15 & 66.67 & 48.65 & 67.74 & 80.15\\
    1   &  0.00   & 71.27 & 48.37 & 0     & 0 & & \\
    }\mydatab

\begin{figure}[ht]
    \begin{subfigure}[t]{1\columnwidth}
        \begin{adjustbox}{width=1\columnwidth}
        \begin{tikzpicture}
            \begin{axis}[
                    x post scale=2,
                    y post scale=0.5,
                    ylabel=Acceptance (\%),
                    xlabel=$\lambda$,
                    ymin=0, ymax=35,
                    xmin=0, xmax=0.9,
                    symbolic x coords={0,0.1,0.2,0.3,0.4,0.5,0.6,0.7,0.8,0.9,1},
                ]
                \addplot [olive, ybar, ybar legend, fill=olive] table[x=Threshold, y=Acceptrate]{\mydatab}; 
            \end{axis}
        \end{tikzpicture}
        \end{adjustbox}
        \caption{Accept rate versus $\lambda$} 
    \end{subfigure}
    
    \begin{subfigure}[t]{1\columnwidth}
        \begin{adjustbox}{width=1\columnwidth}
        \begin{tikzpicture}
            \begin{axis}[
                    legend pos=north west,
                    x post scale=2,
                    y post scale=0.5,
                    ylabel=Recall,
                    xlabel=$\lambda$,
                    ymin=46, ymax=81,
                    xmin=0, xmax=0.9,
                    symbolic x coords={0,0.1,0.2,0.3,0.4,0.5,0.6,0.7,0.8,0.9,1},
                ]
                \addlegendimage{/pgfplots/refstyle=plot_con14_recall_ybtr}\addlegendentry{$y_{BTR}$}
                \addlegendimage{/pgfplots/refstyle=plot_con14_recall_ybase}\addlegendentry{$y_{base}$}
                \addplot+[sharp plot, mark=x, blue] table[x=Threshold,y=Recallforaccept]{\mydatab};\label{plot_con14_recall_ybtr}
                \addplot+[sharp plot, loosely dashed, mark=x, blue] table[x=Threshold,y=Recallybase]{\mydatab};\label{plot_con14_recall_ybase}
            \end{axis}
        \end{tikzpicture}
        \end{adjustbox}
        \caption{Recall for accepted suggestions versus $\lambda$} 
    \end{subfigure}
    
    \begin{subfigure}[t]{1\columnwidth}
        \begin{adjustbox}{width=1\columnwidth}
        \begin{tikzpicture}
            \begin{axis}[
                    legend pos=north west,
                    x post scale=2,
                    y post scale=0.5,
                    ylabel=Precision,
                    xlabel=$\lambda$,
                    ymin=63, ymax=76,
                    xmin=0, xmax=0.9,
                    symbolic x coords={0,0.1,0.2,0.3,0.4,0.5,0.6,0.7,0.8,0.9,1},
                ]
                \addlegendimage{/pgfplots/refstyle=plot_con14_precision_ybtr}\addlegendentry{$y_{BTR}$}
                \addlegendimage{/pgfplots/refstyle=plot_con14_precision_ybase}\addlegendentry{$y_{base}$}
                \addplot+[sharp plot, mark=diamond, red] table[x=Threshold,y=Precisionforaccept]{\mydatab};\label{plot_con14_precision_ybtr}
                \addplot+[sharp plot, loosely dashed, mark=diamond, red] table[x=Threshold,y=Precisionybase]{\mydatab};\label{plot_con14_precision_ybase}
            \end{axis}
        \end{tikzpicture}
        \end{adjustbox}
        \caption{Precision for accepted suggestions versus $\lambda$} 
    \end{subfigure}
    
    \begin{subfigure}[t]{1\columnwidth}
        \begin{adjustbox}{width=1\columnwidth}
        \begin{tikzpicture}
            \begin{axis}[
                    x post scale=2,
                    y post scale=0.5,
                    ylabel=Recall,
                    xlabel=$\lambda$,
                    ymin=48, ymax=49,
                    xmin=0, xmax=1,
                    symbolic x coords={0,0.1,0.2,0.3,0.4,0.5,0.6,0.7,0.8,0.9,1},
                ]
                \addplot+[sharp plot, mark=triangle, black] table[x=Threshold,y=Recall]{\mydatab};
            \end{axis}
        \end{tikzpicture}
        \end{adjustbox}
        \caption{Recall over the whole corpus versus $\lambda$} 
    \end{subfigure}
    
    \begin{subfigure}[t]{1\columnwidth}
        \begin{adjustbox}{width=1\columnwidth}
        \begin{tikzpicture}
            \begin{axis}[
                    x post scale=2,
                    y post scale=0.5,
                    ylabel=Precision,
                    xlabel=$\lambda$,
                    ymin=69, ymax=72,
                    xmin=0, xmax=1,
                    symbolic x coords={0,0.1,0.2,0.3,0.4,0.5,0.6,0.7,0.8,0.9,1},
                ]
                \addplot+[sharp plot, mark=square, black] table[x=Threshold,y=Precision]{\mydatab};      
            \end{axis}
        \end{tikzpicture}
        \end{adjustbox}
        \caption{Precision over the whole corpus versus $\lambda$} 
    \end{subfigure}
    \caption{\textcolor{black}{Precision and recall of BTR ($a_{train}=20, a_{pred}=5$) with respect to different $\lambda$ on the CoNLL-14 set.}} 
    \label{fig:all_versus_lambda_conll14}
\end{figure}
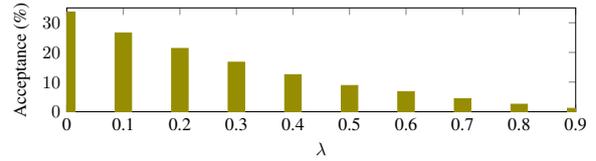
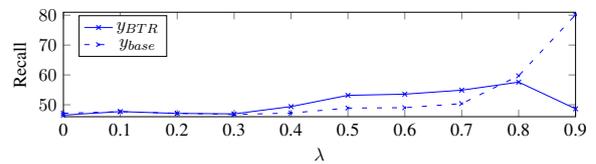
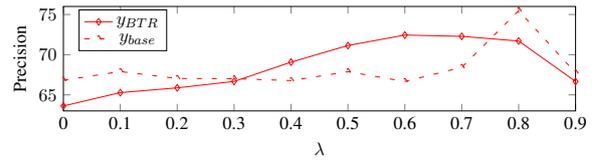
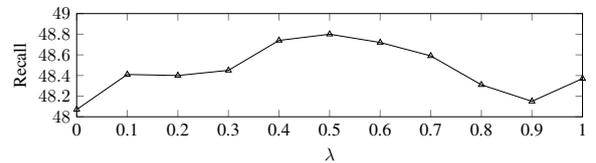
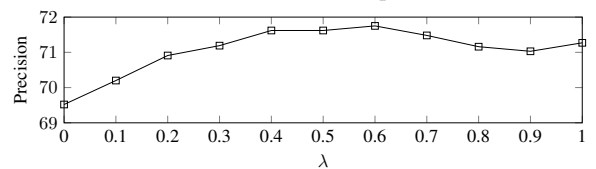

\pgfplotsset{width=7cm,compat=1.15}
\pgfplotstableread[row sep=\\,col sep=&]{
    Threshold & Acceptrate & Precision & Recall & Precisionforaccept & Recallforaccept &Precisionybase & Recallybase \\
	0   & 27.91 & 58.38 & 40.02 & 52.86& 40.30 & 51.84 & 36.02 \\
	0.1 & 19.95 & 59.21 & 39.82 & 55.16& 40.86 & 51.18 & 34.34  \\
	0.2 & 16.18 & 59.42 & 39.66 & 55.74& 40.95 & 50.00 & 33.43 \\
	0.3 & 12.78 & 59.52 & 39.25 & 58.25& 42.45 & 50.34 & 34.43 \\
	0.4 & 10.02 & 59.47 & 39.08 & 58.40& 41.50 & 48.49 & 32.49 \\
	0.5 &  7.67 & 59.40 & 38.76 & 59.27& 41.32 & 47.17 & 32.57 \\
	0.6 &  5.43 & 59.46 & 38.56 & 62.29& 42.58 & 44.80 & 32.81 \\
	0.7 &  3.86 & 59.22 & 38.26 & 62.05& 43.85 & 41.57 & 35.02 \\
	0.8 &  2.78 & 58.96 & 37.98 & 58.67& 44.44 & 39.13 & 40.91 \\
	0.9 &  1.28 & 58.89 & 37.89 & 60.00& 41.67 & 26.55 & 41.67 \\
	1   &  & 58.12 & 37.89 & & & &  \\
    }\mydataallversuslambdabeadev

\begin{figure}[ht]
    \begin{subfigure}[t]{1\columnwidth}
        \begin{adjustbox}{width=1\columnwidth}
        \begin{tikzpicture}
            \begin{axis}[
                    x post scale=2,
                    y post scale=0.5,
                    ylabel=Acceptance (\%),
                    xlabel=$\lambda$,
                    ymin=0, ymax=28,
                    xmin=0, xmax=0.9,
                    symbolic x coords={0,0.1,0.2,0.3,0.4,0.5,0.6,0.7,0.8,0.9,1},
                ]
                \addplot [olive, ybar, ybar legend, fill=olive] table[x=Threshold, y=Acceptrate]{\mydataallversuslambdabeadev}; 
            \end{axis}
        \end{tikzpicture}
        \end{adjustbox}
        \caption{Accept rate versus $\lambda$} 
    \end{subfigure}
    
    \begin{subfigure}[t]{1\columnwidth}
        \begin{adjustbox}{width=1\columnwidth}
        \begin{tikzpicture}
            \begin{axis}[
                    legend pos=south east,
                    x post scale=2,
                    y post scale=0.5,
                    ylabel=Recall,
                    xlabel=$\lambda$,
                    ymin=32, ymax=45,
                    xmin=0, xmax=0.9,
                    symbolic x coords={0,0.1,0.2,0.3,0.4,0.5,0.6,0.7,0.8,0.9,1},
                ]
                \addlegendimage{/pgfplots/refstyle=plot_BEAdev_recall_ybtr}\addlegendentry{$y_{BTR}$}
                \addlegendimage{/pgfplots/refstyle=plot_BEAdev_recall_ybase}\addlegendentry{$y_{base}$}
                \addplot+[sharp plot, mark=x] table[x=Threshold,y=Recallforaccept]{\mydataallversuslambdabeadev};\label{plot_BEAdev_recall_ybtr}
                \addplot+[sharp plot, loosely dashed, mark=x, blue] table[x=Threshold,y=Recallybase]{\mydataallversuslambdabeadev};\label{plot_BEAdev_recall_ybase}
            \end{axis}
        \end{tikzpicture}
        \end{adjustbox}
        \caption{Recall for accepted suggestions versus $\lambda$} 
    \end{subfigure}

    \begin{subfigure}[t]{1\columnwidth}
        \begin{adjustbox}{width=1\columnwidth}
        \begin{tikzpicture}
            \begin{axis}[
                    legend pos=south west,
                    x post scale=2,
                    y post scale=0.5,
                    ylabel=Precision,
                    xlabel=$\lambda$,
                    ymin=26, ymax=63,
                    xmin=0, xmax=0.9,
                    symbolic x coords={0,0.1,0.2,0.3,0.4,0.5,0.6,0.7,0.8,0.9,1},
                ]
                \addlegendimage{/pgfplots/refstyle=plot_BEAdev_precision_ybtr}\addlegendentry{$y_{BTR}$}
                \addlegendimage{/pgfplots/refstyle=plot_BEAdev_precision_ybase}\addlegendentry{$y_{base}$}
                \addplot+[sharp plot, mark=diamond, red] table[x=Threshold,y=Precisionforaccept]{\mydataallversuslambdabeadev};\label{plot_BEAdev_precision_ybtr}
                \addplot+[sharp plot, loosely dashed, mark=diamond, red] table[x=Threshold,y=Precisionybase]{\mydataallversuslambdabeadev};\label{plot_BEAdev_precision_ybase}
            \end{axis}
        \end{tikzpicture}
        \end{adjustbox}
        \caption{Precision for accepted suggestions versus $\lambda$} 
    \end{subfigure}

    \begin{subfigure}[t]{1\columnwidth}
        \begin{adjustbox}{width=1\columnwidth}
        \begin{tikzpicture}
            \begin{axis}[
                    x post scale=2,
                    y post scale=0.5,
                    ylabel=Recall,
                    xlabel=$\lambda$,
                    ymin=37, ymax=41,
                    xmin=0, xmax=1,
                    symbolic x coords={0,0.1,0.2,0.3,0.4,0.5,0.6,0.7,0.8,0.9,1},
                ]
                \addplot+[sharp plot, mark=triangle, black] table[x=Threshold,y=Recall]{\mydataallversuslambdabeadev};
            \end{axis}
        \end{tikzpicture}
        \end{adjustbox}
        \caption{Recall over the whole corpus versus $\lambda$} 
    \end{subfigure}

    \begin{subfigure}[t]{1\columnwidth}
        \begin{adjustbox}{width=1\columnwidth}
        \begin{tikzpicture}
            \begin{axis}[
                    x post scale=2,
                    y post scale=0.5,
                    ylabel=Precision,
                    xlabel=$\lambda$,
                    ymin=58, ymax=60,
                    xmin=0, xmax=1,
                    symbolic x coords={0,0.1,0.2,0.3,0.4,0.5,0.6,0.7,0.8,0.9,1},
                ]
                \addplot+[sharp plot, mark=square, black] table[x=Threshold,y=Precision]{\mydataallversuslambdabeadev};
            \end{axis}
        \end{tikzpicture}
        \end{adjustbox}
        \caption{Precision over the whole corpus versus $\lambda$} 
    \end{subfigure}
    \caption{\textcolor{black}{Precision and recall of BTR ($a_{train}=10, a_{pred}=5$) with respect to different $\lambda$ on the BEA dev set.}} 
    \label{fig:all_versus_lambda_beadev}
\end{figure}
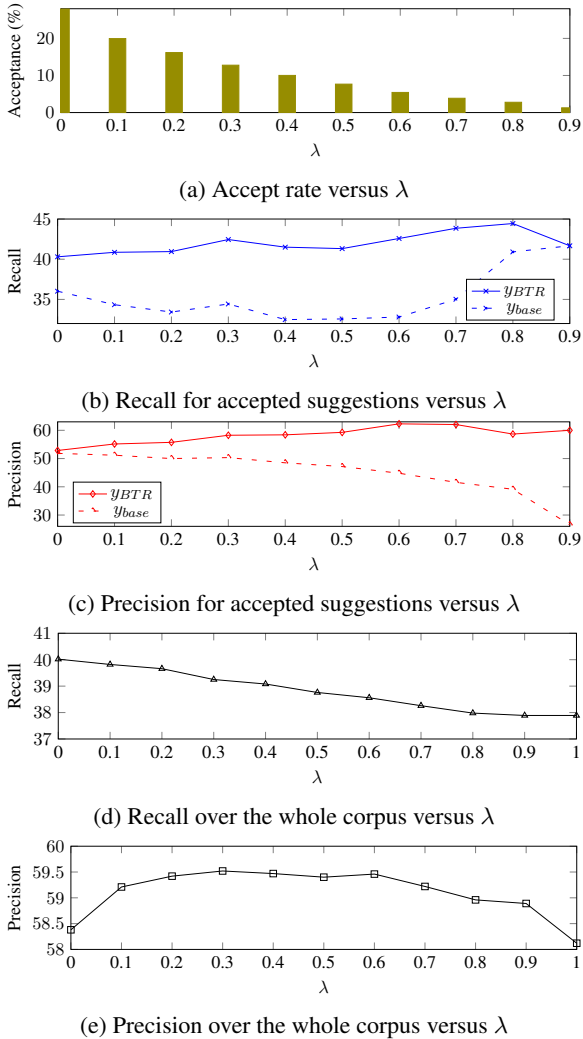

\pgfplotsset{width=7cm,compat=1.15}
\pgfplotstableread[row sep=\\,col sep=&]{
    Threshold & Acceptrate & Precision & Recall \\
	0   & 26.74 & 72.45 & 61.10 \\
	0.1 & 20.08 & 73.56 & 60.96 \\
	0.2 & 16.02 & 74.16 & 61.11 \\
	0.3 & 12.82 & 74.26 & 60.85 \\
	0.4 & 9.67 & 74.33 & 60.44 \\
	0.5 & 7.24 & 74.24 & 60.11 \\
	0.6 & 5.00 & 74.35 & 59.74 \\
	0.7 & 3.19 & 74.26 & 59.66 \\
	0.8 & 1.83 & 74.30 & 59.53 \\
	0.9 & 0.92 & 74.27 & 59.54 \\
	1   & 0 &  73.96 & 59.45 \\
    }\mydataallversuslambdabeatest

\begin{figure}[ht]
    \begin{subfigure}[t]{1\columnwidth}
        \begin{adjustbox}{width=1\columnwidth}
        \begin{tikzpicture}
            \begin{axis}[
                    x post scale=2,
                    y post scale=0.5,
                    ylabel=Acceptance (\%),
                    xlabel=$\lambda$,
                    ymin=0, ymax=27,
                    xmin=0, xmax=0.9,
                    symbolic x coords={0,0.1,0.2,0.3,0.4,0.5,0.6,0.7,0.8,0.9,1},
                ]
                \addplot [olive, ybar, ybar legend, fill=olive] table[x=Threshold, y=Acceptrate]{\mydataallversuslambdabeatest}; 
            \end{axis}
        \end{tikzpicture}
        \end{adjustbox}
        \caption{Accept rate versus $\lambda$} 
    \end{subfigure}

    \begin{subfigure}[t]{1\columnwidth}
        \begin{adjustbox}{width=1\columnwidth}
        \begin{tikzpicture}
            \begin{axis}[
                    x post scale=2,
                    y post scale=0.5,
                    ylabel=Recall,
                    xlabel=$\lambda$,
                    ymin=59, ymax=62,
                    xmin=0, xmax=1,
                    symbolic x coords={0,0.1,0.2,0.3,0.4,0.5,0.6,0.7,0.8,0.9,1},
                ]
                \addplot+[sharp plot, mark=triangle] table[x=Threshold,y=Recall]{\mydataallversuslambdabeatest};
            \end{axis}
        \end{tikzpicture}
        \end{adjustbox}
        \caption{Recall over the whole corpus versus $\lambda$} 
    \end{subfigure}

    \begin{subfigure}[t]{1\columnwidth}
        \begin{adjustbox}{width=1\columnwidth}
        \begin{tikzpicture}
            \begin{axis}[
                    x post scale=2,
                    y post scale=0.5,
                    ylabel=Precision,
                    xlabel=$\lambda$,
                    ymin=72, ymax=75,
                    xmin=0, xmax=1,
                    symbolic x coords={0,0.1,0.2,0.3,0.4,0.5,0.6,0.7,0.8,0.9,1},
                ]
                \addplot+[sharp plot, mark=square, red] table[x=Threshold,y=Precision]{\mydataallversuslambdabeatest};
            \end{axis}
        \end{tikzpicture}
        \end{adjustbox}
        \caption{Precision over the whole corpus versus $\lambda$} 
    \end{subfigure}
    \caption{\textcolor{black}{Precision and recall of BTR ($a_{train}=10, a_{pred}=5$) with respect to different $\lambda$ on the BEA test set.}} 
    \label{fig:all_versus_lambda_beatest}
\end{figure}
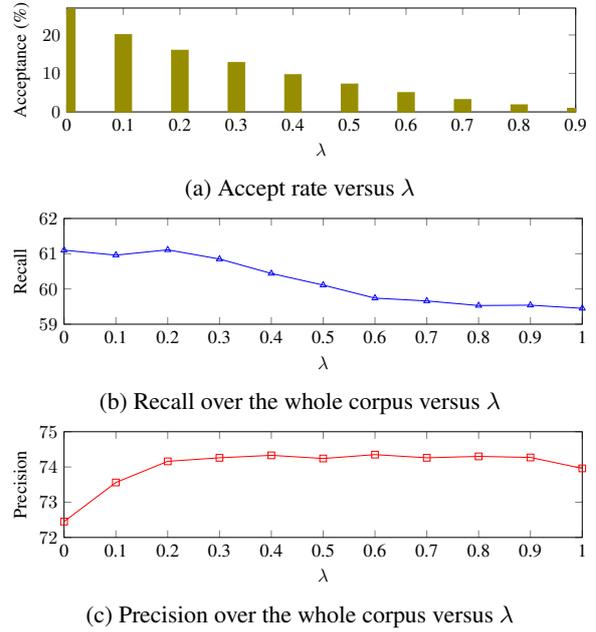

\pgfplotsset{width=7cm,compat=1.15}
\pgfplotstableread[row sep=\\,col sep=&]{
	Position & BTRatrain0 & BTRatrain5 & BTRatrain10 & BTRatrain20 & R2L & RoBERTaatrain0 & RoBERTaatrain5 & RoBERTaatrain10 & RoBERTaatrain20 \\
	1 & 0.2391 & 0.6749 & 0.7057 & 0.7464 & 0.3251 & 0.2333 & 0.244 & 0.2420 & 0.2409 \\
	2 & 0.2181 & 0.1658 & 0.1642 & 0.1433 & 0.2083 & 0.2182 & 0.2193 & 0.2190 & 0.2188 \\
	3 & 0.1996 & 0.0787 & 0.067 & 0.0587 & 0.1734 & 0.2001 & 0.1979 & 0.1981 & 0.1982 \\
	4 & 0.182 & 0.0498 & 0.0399 & 0.0332 & 0.1565 & 0.1841 & 0.1805 & 0.1811 & 0.1814 \\
	5 & 0.1612 & 0.0309 & 0.0232 & 0.0185 & 0.1366 & 0.1642 & 0.1583 & 0.1599 & 0.1607 \\
    }\mydatarankdistributionBEA

\begin{figure}[t!]
    \begin{adjustbox}{width=1\columnwidth}
        \begin{tikzpicture}
            \begin{axis}[
                scale only axis,
                legend pos=outer north east,
                legend style={nodes={scale=0.6, transform shape}},
                x post scale=1,
                y post scale=1,
                ylabel=Probability,
                xlabel=Ranking,
                ymin=0, ymax=0.8,
                xmin=1, xmax=5,
            ]
            \addlegendimage{/pgfplots/refstyle=plot_BTRatrain0_ranking_BEA}\addlegendentry{$BTR, a_{train}=0$}
            \addlegendimage{/pgfplots/refstyle=plot_BTRatrain5_ranking_BEA}\addlegendentry{$BTR, a_{train}=5$}
            \addlegendimage{/pgfplots/refstyle=plot_BTRatrain10_ranking_BEA}\addlegendentry{$BTR, a_{train}=10$}
            \addlegendimage{/pgfplots/refstyle=plot_BTRatrain20_ranking_BEA}\addlegendentry{$BTR, a_{train}=20$}
            \addlegendimage{/pgfplots/refstyle=plot_R2L_ranking_BEA}\addlegendentry{$R2L$}
            \addlegendimage{/pgfplots/refstyle=plot_RoBERTaatrain0_ranking_BEA}\addlegendentry{$RoBERTa, a_{train}=0$}
            \addlegendimage{/pgfplots/refstyle=plot_RoBERTaatrain5_ranking_BEA}\addlegendentry{$RoBERTa, a_{train}=5$}
            \addlegendimage{/pgfplots/refstyle=plot_RoBERTaatrain10_ranking_BEA}\addlegendentry{$RoBERTa, a_{train}=10$}
            \addlegendimage{/pgfplots/refstyle=plot_RoBERTaatrain20_ranking_BEA}\addlegendentry{$RoBERTa, a_{train}=20$}
            \addplot+[sharp plot, mark=none, red] table[x=Position,y=BTRatrain0]{\mydatarankdistributionBEA};\label{plot_BTRatrain0_ranking_BEA}
            \addplot+[sharp plot, mark=none, blue, dashed] table[x=Position,y=BTRatrain5]{\mydatarankdistributionBEA};\label{plot_BTRatrain5_ranking_BEA}
            \addplot+[sharp plot, mark=none, blue, dash dot] table[x=Position,y=BTRatrain10]{\mydatarankdistributionBEA};\label{plot_BTRatrain10_ranking_BEA}
            \addplot+[sharp plot, mark=none, blue] table[x=Position,y=BTRatrain20]{\mydatarankdistributionBEA};\label{plot_BTRatrain20_ranking_BEA}
            \addplot+[sharp plot, mark=none, black] table[x=Position,y=R2L]{\mydatarankdistributionBEA};\label{plot_R2L_ranking_BEA}
            \addplot+[sharp plot, mark=none, black] table[x=Position,y=RoBERTaatrain0]{\mydatarankdistributionBEA};\label{plot_RoBERTaatrain0_ranking_BEA}
            \addplot+[sharp plot, mark=none, black] table[x=Position,y=RoBERTaatrain5]{\mydatarankdistributionBEA};\label{plot_RoBERTaatrain5_ranking_BEA}
            \addplot+[sharp plot, mark=none, black] table[x=Position,y=RoBERTaatrain10]{\mydatarankdistributionBEA};\label{plot_RoBERTaatrain10_ranking_BEA}
            \addplot+[sharp plot, mark=none, black] table[x=Position,y=RoBERTaatrain20]{\mydatarankdistributionBEA};\label{plot_RoBERTaatrain20_ranking_BEA}
        \end{axis}
        \end{tikzpicture}
    \end{adjustbox}
\caption{\textcolor{black}{Average probability for each rank on the BEA test set. The top-5 candidate sentences were generated by T5GEC.}}
\label{fig:probability_distribution_of_ranking_BEA}
\end{figure}
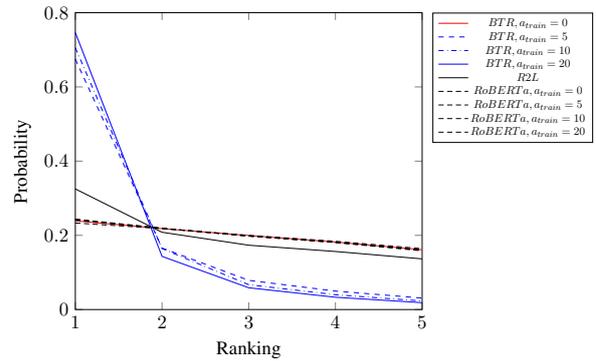

\pgfplotsset{width=7cm,compat=1.15}
\pgfplotstableread[row sep=\\,col sep=&]{
    Threshold & Acceptrate & GLEU  & GLEUforaccept &GLEUybase\\
	0 & 51.27 & 60.14 & 64.21 & 64.55  \\
	0.1 & 24.63 & 60.37 & 66.81 & 65.81 \\
	0.2 & 13.12 & 59.98 & 69.98 & 68.20 \\
	0.3 & 6.96 & 59.82 & 68.48 & 65.54 \\
	0.4 & 3.88 & 59.67 & 69.47 & 64.62 \\
	0.5 & 1.47 & 59.68 & 67.49 & 55.98 \\
	0.6 & 0.94 & 59.65 & 65.93 & 51.09 \\
	0.7 & 0.80 & 59.64 & 65.68 & 50.53 \\
	0.8 & 0.40 & 59.63 & 51.97 & 34.54 \\
	0.9 & 0.40 & 59.63 & 51.97 & 34.54 \\
	1 & & 59.26 & & \\
    }\mydatagleuversuslambda

\pgfplotsset{width=7cm,compat=1.15}
\pgfplotstableread[row sep=\\,col sep=&]{
    Threshold & Acceptrate & GLEU  & GLEUforaccept &GLEUybase\\
	0   & 49.73 & 54.46 & 58.63 & 58.97 \\
	0.1 & 26.53 & 54.44 & 63.50 & 63.20 \\
	0.2 & 10.88 & 53.88 & 65.67 & 64.45 \\
	0.3 & 5.31  & 53.43 & 69.32 & 67.87 \\
	0.4 & 3.71 & 53.33 & 70.07 & 69.17 \\
	0.5 & 2.12 & 53.29 & 69.62 & 68.37 \\
	0.6 & 1.46 & 53.22 & 68.24 & 67.81 \\
	0.7 & 1.06 & 53.22 & 72.48 & 72.49 \\
	0.8 & 0.93 & 53.26 & 72.63 & 71.71 \\
	0.9 & 0.40 & 53.20 & 60.01 & 64.16 \\
	1   &  & 53.25 & & \\
    }\mydatagleuversuslambdadev

\begin{figure}[ht]
    \begin{subfigure}[t]{1\columnwidth}
        \begin{adjustbox}{width=1\columnwidth}
        \begin{tikzpicture}
            \begin{axis}[
                    x post scale=2,
                    y post scale=0.5,
                    ylabel=Acceptance (\%),
                    xlabel=$\lambda$,
                    ymin=0, ymax=50,
                    xmin=0, xmax=0.9,
                    symbolic x coords={0,0.1,0.2,0.3,0.4,0.5,0.6,0.7,0.8,0.9,1},
                ]
                \addplot [olive, ybar, ybar legend, fill=olive] table[x=Threshold, y=Acceptrate]{\mydatagleuversuslambdadev}; 
            \end{axis}
        \end{tikzpicture}
        \end{adjustbox}
        \caption{Accept rate versus $\lambda$} 
    \end{subfigure}
    
    \begin{subfigure}[t]{1\columnwidth}
        \begin{adjustbox}{width=1\columnwidth}
        \begin{tikzpicture}
            \begin{axis}[
                    legend pos=north west,
                    x post scale=2,
                    y post scale=0.5,
                    ylabel=GLEU,
                    xlabel=$\lambda$,
                    ymin=58, ymax=73,
                    xmin=0, xmax=0.9,
                    symbolic x coords={0,0.1,0.2,0.3,0.4,0.5,0.6,0.7,0.8,0.9,1},
                ]
                \addlegendimage{/pgfplots/refstyle=plot_JFLEGdev_GLEU_ybtr}\addlegendentry{$y_{BTR}$}
                \addlegendimage{/pgfplots/refstyle=plot_JFLEGdev_GLEU_ybase}\addlegendentry{$y_{base}$}
                \addplot+[sharp plot, mark=x, blue] table[x=Threshold,y=GLEUforaccept]{\mydatagleuversuslambdadev};\label{plot_JFLEGdev_GLEU_ybtr}
                \addplot+[sharp plot, loosely dashed, mark=x, blue] table[x=Threshold,y=GLEUybase]{\mydatagleuversuslambdadev};\label{plot_JFLEGdev_GLEU_ybase}
            \end{axis}
        \end{tikzpicture}
        \end{adjustbox}
        \caption{GLEU for accepted suggestions versus $\lambda$} 
    \end{subfigure}

    \begin{subfigure}[t]{1\columnwidth}
        \begin{adjustbox}{width=1\columnwidth}
        \begin{tikzpicture}
            \begin{axis}[
                    x post scale=2,
                    y post scale=0.5,
                    ylabel=GLEU,
                    xlabel=$\lambda$,
                    ymin=53, ymax=55,
                    xmin=0, xmax=1,
                    symbolic x coords={0,0.1,0.2,0.3,0.4,0.5,0.6,0.7,0.8,0.9,1},
                ]
                \addplot+[sharp plot, mark=triangle, black] table[x=Threshold,y=GLEU]{\mydatagleuversuslambdadev};
            \end{axis}
        \end{tikzpicture}
        \end{adjustbox}
        \caption{GLEU over the whole corpus versus $\lambda$} 
    \end{subfigure}
    \caption{\textcolor{black}{GLEU of BTR ($a_{train}=5, a_{pred}=15$) with respect to different $\lambda$ on the JFLEG dev set.}}
    \label{fig:all_versus_lambda_jflegdev}
\end{figure}
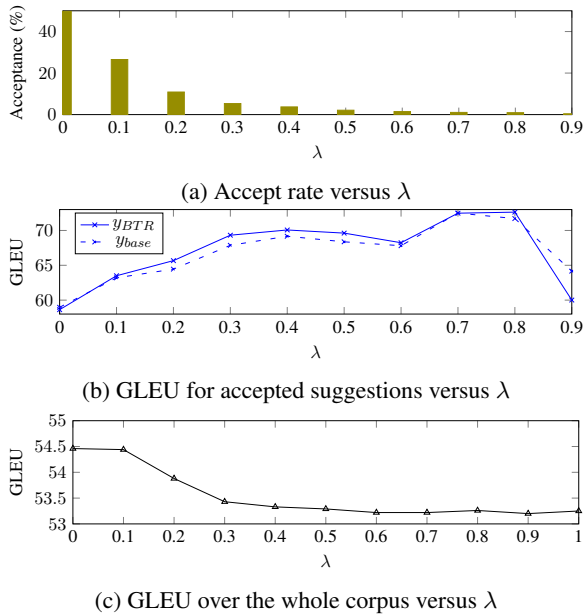

\begin{figure}[ht]
    \begin{subfigure}[t]{1\columnwidth}
        \begin{adjustbox}{width=1\columnwidth}
        \begin{tikzpicture}
            \begin{axis}[
                    x post scale=2,
                    y post scale=0.5,
                    ylabel=Acceptance (\%),
                    xlabel=$\lambda$,
                    ymin=0, ymax=52,
                    xmin=0, xmax=0.9,
                    symbolic x coords={0,0.1,0.2,0.3,0.4,0.5,0.6,0.7,0.8,0.9,1},
                ]
                \addplot [olive, ybar, ybar legend, fill=olive] table[x=Threshold, y=Acceptrate]{\mydatagleuversuslambda}; 
            \end{axis}
        \end{tikzpicture}
        \end{adjustbox}
        \caption{Accept rate versus $\lambda$} 
    \end{subfigure}
    
    \begin{subfigure}[t]{1\columnwidth}
        \begin{adjustbox}{width=1\columnwidth}
        \begin{tikzpicture}
            \begin{axis}[
                    legend pos=south west,
                    x post scale=2,
                    y post scale=0.5,
                    ylabel=GLEU,
                    xlabel=$\lambda$,
                    ymin=34, ymax=70,
                    xmin=0, xmax=0.9,
                    symbolic x coords={0,0.1,0.2,0.3,0.4,0.5,0.6,0.7,0.8,0.9,1},
                ]
                \addlegendimage{/pgfplots/refstyle=plot_JFLEGtest_GLEU_ybtr}\addlegendentry{$y_{BTR}$}
                \addlegendimage{/pgfplots/refstyle=plot_JFLEGtest_GLEU_ybase}\addlegendentry{$y_{base}$}
                \addplot+[sharp plot, mark=x, blue] table[x=Threshold,y=GLEUforaccept]{\mydatagleuversuslambda};\label{plot_JFLEGtest_GLEU_ybtr}
                \addplot+[sharp plot, loosely dashed, mark=x, blue] table[x=Threshold,y=GLEUybase]{\mydatagleuversuslambda};\label{plot_JFLEGtest_GLEU_ybase}
            \end{axis}
        \end{tikzpicture}
        \end{adjustbox}
        \caption{GLEU for accepted suggestions versus $\lambda$} 
    \end{subfigure}

    \begin{subfigure}[t]{1\columnwidth}
        \begin{adjustbox}{width=1\columnwidth}
        \begin{tikzpicture}
            \begin{axis}[
                    x post scale=2,
                    y post scale=0.5,
                    ylabel=GLEU,
                    xlabel=$\lambda$,
                    ymin=59, ymax=61,
                    xmin=0, xmax=1,
                    symbolic x coords={0,0.1,0.2,0.3,0.4,0.5,0.6,0.7,0.8,0.9,1},
                ]
                \addplot+[sharp plot, mark=triangle, black] table[x=Threshold,y=GLEU]{\mydatagleuversuslambda};
            \end{axis}
        \end{tikzpicture}
        \end{adjustbox}
        \caption{GLEU over the whole corpus versus $\lambda$} 
    \end{subfigure}
    \caption{\textcolor{black}{GLEU of BTR ($a_{train}=5, a_{pred}=15$) with respect to different $\lambda$ on the JFLEG test set.}} 
    \label{fig:all_versus_lambda_jflegtest}
\end{figure}
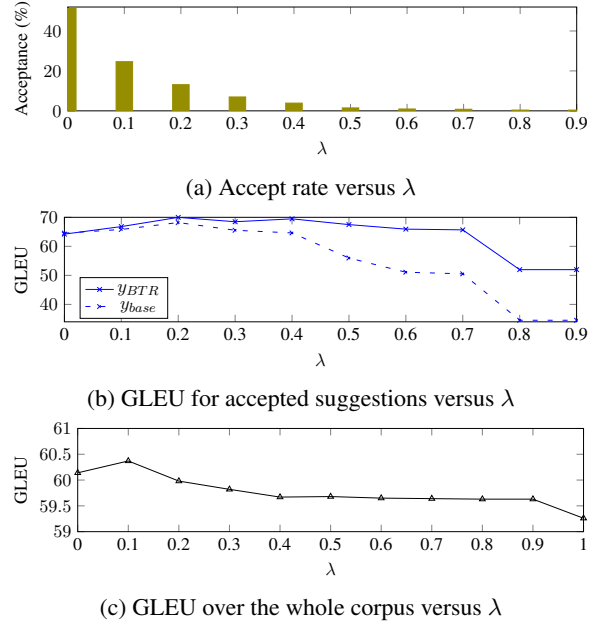

\pgfplotsset{width=7cm,compat=1.15}
\pgfplotstableread[row sep=\\,col sep=&]{
	Position & BTRatrain0 & BTRatrain5 & BTRatrain10 & BTRatrain20 & R2L & RoBERTaatrain0 & RoBERTaatrain5 & RoBERTaatrain10 & RoBERTaatrain20 \\
	1 & 0.0783 & 0.3725 & 0.4388 & 0.4604 & 0.1259 & 0.0848 & 0.0965 & 0.0940 & 0.0884 \\
	2 & 0.0767 & 0.1530 & 0.1662 & 0.1691 & 0.0941 & 0.0807 & 0.0843 & 0.0832 & 0.0823 \\
	3 & 0.0752 & 0.0896 & 0.0872 & 0.0867 & 0.0789 & 0.0772 & 0.0782 & 0.0778 & 0.0781 \\
	4 & 0.0741 & 0.0700 & 0.0637 & 0.0618 & 0.0746 & 0.0755 & 0.0760 & 0.0757 & 0.0760 \\
	5 & 0.0725 & 0.0568 & 0.0486 & 0.0457 & 0.0704 & 0.0733 & 0.0734 & 0.0732 & 0.0736 \\
	6 & 0.0712 & 0.0481 & 0.0393 & 0.0368 & 0.0675 & 0.0712 & 0.0706 & 0.0708 & 0.0712 \\
	7 & 0.0694 & 0.0411 & 0.0325 & 0.0299 & 0.0645 & 0.0689 & 0.0680 & 0.0683 & 0.0688 \\
	8 & 0.0676 & 0.0356 & 0.0277 & 0.0253 & 0.0622 & 0.0671 & 0.0659 & 0.0663 & 0.0667 \\
	9 & 0.0657 & 0.0302 & 0.0228 & 0.0206 & 0.0594 & 0.0647 & 0.0633 & 0.0637 & 0.0643 \\
	10 & 0.0638 & 0.0259 & 0.0191 & 0.0171 & 0.0569 & 0.0625 & 0.0607 & 0.0612 & 0.0618 \\
	11 & 0.0620 & 0.0222 & 0.0163 & 0.0143 & 0.0548 & 0.0603 & 0.0584 & 0.0590 & 0.0596 \\
	12 & 0.0598 & 0.0184 & 0.0130 & 0.0113 & 0.0522 & 0.0579 & 0.0559 & 0.0564 & 0.0570 \\
	13 & 0.0577 & 0.0154 & 0.0108 & 0.0091 & 0.0496 & 0.0555 & 0.0531 & 0.0537 & 0.0544 \\
	14 & 0.0549 & 0.0121 & 0.0081 & 0.0068 & 0.0465 & 0.0523 & 0.0500 & 0.0505 & 0.0511 \\
	15 & 0.0501 & 0.0092 & 0.0060 & 0.0050 & 0.0424 & 0.0483 & 0.0457 & 0.0463 & 0.0468 \\
    }\mydatarankdistributionJFLEG

\begin{figure}[t]
    \begin{adjustbox}{width=1\columnwidth}
        \begin{tikzpicture}
            \begin{axis}[
                scale only axis,
                legend pos=outer north east,
                legend style={nodes={scale=0.6, transform shape}},
                x post scale=1,
                y post scale=1,
                ylabel=Probability,
                xlabel=Ranking,
                ymin=0, ymax=0.5,
                xmin=1, xmax=15,
            ]
            \addlegendimage{/pgfplots/refstyle=plot_BTRatrain0_ranking_JFLEG}\addlegendentry{$BTR, a_{train}=0$}
            \addlegendimage{/pgfplots/refstyle=plot_BTRatrain5_ranking_JFLEG}\addlegendentry{$BTR, a_{train}=5$}
            \addlegendimage{/pgfplots/refstyle=plot_BTRatrain10_ranking_JFLEG}\addlegendentry{$BTR, a_{train}=10$}
            \addlegendimage{/pgfplots/refstyle=plot_BTRatrain20_ranking_JFLEG}\addlegendentry{$BTR, a_{train}=20$}
            \addlegendimage{/pgfplots/refstyle=plot_R2L_ranking_JFLEG}\addlegendentry{$R2L$}
            \addlegendimage{/pgfplots/refstyle=plot_RoBERTaatrain0_ranking_JFLEG}\addlegendentry{$RoBERTa, a_{train}=0$}
            \addlegendimage{/pgfplots/refstyle=plot_RoBERTaatrain5_ranking_JFLEG}\addlegendentry{$RoBERTa, a_{train}=5$}
            \addlegendimage{/pgfplots/refstyle=plot_RoBERTaatrain10_ranking_JFLEG}\addlegendentry{$RoBERTa, a_{train}=10$}
            \addlegendimage{/pgfplots/refstyle=plot_RoBERTaatrain20_ranking_JFLEG}\addlegendentry{$RoBERTa, a_{train}=20$}
            \addplot+[sharp plot, mark=none, red] table[x=Position,y=BTRatrain0]{\mydatarankdistributionJFLEG};\label{plot_BTRatrain0_ranking_JFLEG}
            \addplot+[sharp plot, mark=none, blue, dashed] table[x=Position,y=BTRatrain5]{\mydatarankdistributionJFLEG};\label{plot_BTRatrain5_ranking_JFLEG}
            \addplot+[sharp plot, mark=none, blue, dash dot] table[x=Position,y=BTRatrain10]{\mydatarankdistributionJFLEG};\label{plot_BTRatrain10_ranking_JFLEG}
            \addplot+[sharp plot, mark=none, blue] table[x=Position,y=BTRatrain20]{\mydatarankdistributionJFLEG};\label{plot_BTRatrain20_ranking_JFLEG}
            \addplot+[sharp plot, mark=none, black] table[x=Position,y=R2L]{\mydatarankdistributionJFLEG};\label{plot_R2L_ranking_JFLEG}
            \addplot+[sharp plot, mark=none, black] table[x=Position,y=RoBERTaatrain0]{\mydatarankdistributionJFLEG};\label{plot_RoBERTaatrain0_ranking_JFLEG}
            \addplot+[sharp plot, mark=none, black] table[x=Position,y=RoBERTaatrain5]{\mydatarankdistributionJFLEG};\label{plot_RoBERTaatrain5_ranking_JFLEG}
            \addplot+[sharp plot, mark=none, black] table[x=Position,y=RoBERTaatrain10]{\mydatarankdistributionJFLEG};\label{plot_RoBERTaatrain10_ranking_JFLEG}
            \addplot+[sharp plot, mark=none, black] table[x=Position,y=RoBERTaatrain20]{\mydatarankdistributionJFLEG};\label{plot_RoBERTaatrain20_ranking_JFLEG}
        \end{axis}
        \end{tikzpicture}
    \end{adjustbox}
\caption{\textcolor{black}{Average probability for each rank on the JFLEG test set. The top-15 candidate sentences were generated by T5GEC.}}
\label{fig:probability_distribution_of_ranking_JFLEG}
\end{figure}
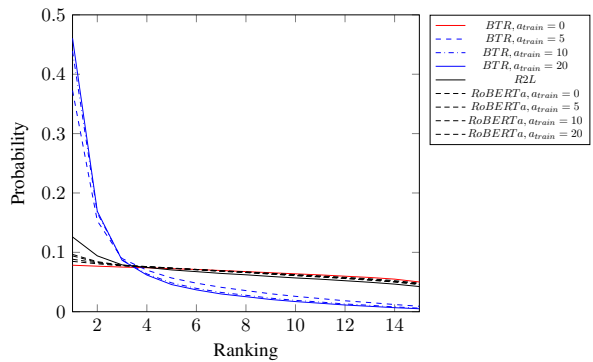

\section{Precision and Recall With T5GEC (large) Candidates}
\label{sec:appendix_precision_recall}
\begin{table*}[ht]
\centering
\resizebox{0.7\linewidth}{!}{
\begin{tabular}{lllllllll}
\toprule
\multirow{2}{*}{\textbf{Model}}  & \multicolumn{2}{c}{\textbf{CoNLL-13}} & \multicolumn{2}{c}{\textbf{CoNLL-14}} & \multicolumn{2}{c}{\textbf{BEA test}}  \\ \cmidrule(lr){2-3} \cmidrule(lr){4-5} \cmidrule(lr){6-7}
  & \multicolumn{1}{c}{\textbf{p}} & \multicolumn{1}{c}{\textbf{r}} & \multicolumn{1}{c}{\textbf{p}} & \multicolumn{1}{c}{\textbf{r}} & \multicolumn{1}{c}{\textbf{p}} & \multicolumn{1}{c}{\textbf{r}}  \\ \midrule
Oracle & 66.34 & 35.34 & 76.04 & 53.36 & - & - \\ \midrule
T5GEC (large) & 60.24 & 31.20 & 73.10 & \textbf{49.76} & 75.65 & 60.87 \\
R2L & \textbf{61.55} & 30.03 & \textbf{73.60} & 48.47 & \textbf{77.06} & 60.24 \\
RoBERTa ($\lambda=0.1$) w/o $a_{train}$ & 60.22 & 31.17 & 73.12 & \textbf{49.76} & 75.74 & 60.83 \\ \midrule
BTR ($\lambda=0.8$) & 60.54 & \textbf{31.28} & 72.71 & \textbf{49.76} & 75.91 & \textbf{61.13} \\ \bottomrule
\end{tabular}
}
\caption{The (p)recision and (r)ecall on each dataset. The top-5 candidate sentences were generated by T5GEC (large). Bold scores represent the highest precision and recall for each dataset.}
\label{table:precision_recall_detail_large}
\end{table*}

Given the top-5 candidate sentences generated by T5GEC (large), we compared the precision and recall of the BTR with those of R2L and RoBERTa in Table \ref{table:precision_recall_detail_large}.

\section{Example of Reranked Outputs}
\label{sec:appendix_example_of_reranked_outputs}

Table \ref{table:example_of_reranked_outputs} provides examples of ranked outputs by T5GEC, R2L, RoBERTa w/o $a_{train}$ ($\lambda=0.1$), and BTR ($\lambda=0.4$). 
\textcolor{black}{The first block of output results demonstrates the difficulty of correcting spelling errors.}
In this block, the BTR outputs the token ``insensitively'' with the correct spelling but a mismatched meaning, whereas other rerankers tend to keep the original token ``intesively'' with a spelling error. 
\textcolor{black}{The examples in the second block show that both the BTR and R2L are capable of correctly addressing verb tense errors.}
The examples in the last block show that even though the BTR recognizes the missing determiner ``the'' for the word ``Disadvantage'', it still misses a that-clause sentence.

\begin{table*}[ht]
\resizebox{1\linewidth}{!}{
\begin{tabular}{ll}
\hline
Source & However , it is \textcolor{blue}{a} good practice not to \textcolor{blue}{intesively} use social media all the time .   \\
Gold 1 & However , it is \textcolor{blue}{a} good practice not to \textcolor{blue}{intensely} use social media all the time .   \\
Gold 2 & However , it is good practice not to \textcolor{blue}{intensively} use social media all the time .  \\
Candidate 1 (R2L, RoBERTa, T5GEC) & However , it is good practice not to \textcolor{blue}{intesively} use social media all the time .  \\
Candidate 2 & However , it is good practice not to \textcolor{blue}{intensely} use social media all the time .  \\
Candidate 3 (BTR) & However , it is good practice not to \textcolor{blue}{insensitively} use social media all the time .  \\ \hline
Source & It is true that social media \textcolor{blue}{makes} people \textcolor{blue}{be able to connect} one another more conveniently . \\
Gold 1 & It is true that social media \textcolor{blue}{allows} people \textcolor{blue}{to connect to} one another more conveniently .  \\
Gold 2 & It is true that social media \textcolor{blue}{make} people \textcolor{blue}{able to connect with} one another more conveniently   \\
Candidate 1 (RoBERTa, T5GEC) & It is true that social media \textcolor{blue}{makes} people \textcolor{blue}{be able to connect with} one another more conveniently . \\
Candidate 2 (BTR, R2L) & It is true that social media \textcolor{blue}{makes} people \textcolor{blue}{able to connect with} one another more conveniently . \\
Candidate 3 & It is true that social media \textcolor{blue}{makes} people \textcolor{blue}{able to connect to} one another more conveniently . \\ \hline
Source & \textcolor{blue}{Disadvantage is} parking their \textcolor{blue}{car} is very difficult .  \\
Gold 1 & \textcolor{blue}{A disadvantage is that} parking their \textcolor{blue}{cars} is very difficult . \\
Gold 2 & \textcolor{blue}{A disadvantage is that} parking their \textcolor{blue}{car} is very difficult . \\
Gold 3 & \textcolor{blue}{The disadvantage is that} parking their \textcolor{blue}{car} is very difficult . \\
Candidate 1 (R2L, RoBERTa, T5GEC) & \textcolor{blue}{Disadvantage is} parking their \textcolor{blue}{car} is very difficult .  \\
Candidate 2 (BTR) & \textcolor{blue}{The disadvantage is} parking their \textcolor{blue}{car} is very difficult .  \\
Candidate 3 & \textcolor{blue}{The disadvantage is that} parking their \textcolor{blue}{car} is very difficult . \\ \hline
\end{tabular}
}
\caption{Examples of reranked outputs. The 3 candidate sentences were generated by T5GEC. \textcolor{blue}{Blue} indicates the range of corrections. 
Examples in the first two and last block were extracted from the CoNLL-14 and JFLEG test corpus, respectively.}
\label{table:example_of_reranked_outputs}
\end{table*}

\section{Inference Time Cost}
\begin{table*}[ht]
\centering
\begin{tabular}{lrrrrrr}
\toprule
\textbf{Model} & \textbf{CoNLL-13} & \textbf{CoNLL-14} & \textbf{BEA dev} & \textbf{BEA test} & \textbf{JFLEG dev} & \textbf{JFLEG test} \\ \midrule
T5GEC & 778 & 790 & 3638 & 3776 & 451 & 444 \\
BERT & 22 & 21 & 68 & 69 & 12 & 13 \\
R2L & 34 & 32 & 108 & 109 & 19 & 19 \\
RoBERTa & 82 & 88 & 333 & 386 & 46 & 69 \\
BTR & 194 & 199 & 740 & 738 & 113 & 122 \\ \bottomrule
\end{tabular}
\caption{Time cost (seconds) in inference over the whole corpus with 5 candidates generated by T5GEC.}
\label{table:inference_time_cost}
\end{table*}

In inference, we required all rerankers to compute one target sequence at a time to estimate the time cost. For RoBERTa and the BTR, we rearranged the given target sequence by masking each token. 
\textcolor{black}{These rearranged sequences were then put into a mini-batch for parallel computation.}
For T5GEC, given the source sentence, we used the mini-batch with a size of 5 to parallelly compute all beams.

Table \ref{table:inference_time_cost} displays the time cost for each model to estimate scores over the entire corpus with 5 candidates, using one NVIDIA A100 80GB GPU. We only calculated the time for estimating probability and ignored the time for loading the model and dataset. T5GEC costs the most time among all rerankers, as it predicts tokens of the target sequence one by one. RoBERTa and the BTR took longer than BERT and R2L due to the target sequence rearrangement procedure. The BTR took 2 to 3 times as much as RoBERTa due to the additional decoder structure.